\pdfoutput=1

\documentclass[11pt]{article}
\DeclareUnicodeCharacter{2223}{\mid}

\usepackage{acl}

\usepackage{pifont}
\usepackage{amsmath} 
\usepackage{float} 
\usepackage{color}
\usepackage{amsfonts}
\usepackage{bm}
\usepackage{placeins}
\usepackage{multirow}
\usepackage{subcaption}
\usepackage{booktabs}
\usepackage{enumitem}
\usepackage{adjustbox}
\usepackage{caption}
\usepackage{array}
\usepackage{tabularx}
\usepackage{amssymb}
\usepackage{times}
\usepackage{latexsym}
\usepackage[T1]{fontenc}
\usepackage[utf8]{inputenc}
\usepackage{microtype}
\usepackage{inconsolata}
\usepackage{graphicx}
\usepackage{hhline}
\usepackage{colortbl}

\renewcommand{\small}{\fontsize{9.5pt}{9.5pt}\selectfont}

\title{Improving Reasoning Capabilities in Small Models through Mixture-of-Layers Distillation with Stepwise Attention on Key Information}

\author{
    Yao Chen$^{1, 2}$, Jiawei Sheng$^{1}$, Wenyuan Zhang$^{1, 2}$, Tingwen Liu$^{1, 2}$\thanks{indicates corresponding author.}\\
  $^1$Institute of Information Engineering, Chinese Academy of Sciences \\
  $^2$School of Cyber Security, University of Chinese Academy of Sciences \\
  \texttt{\{chenyao2023, shengjiawei, zhangwenyuan, liutingwen\}@iie.ac.cn}
}

\begin{document}
\maketitle
\begin{abstract}
The significant computational demands of large language models have increased interest in distilling reasoning abilities into smaller models via Chain-of-Thought (CoT) distillation.
Current CoT distillation methods mainly focus on transferring teacher-generated rationales for complex reasoning to student models.
However, they do not adequately explore teachers' dynamic attention toward critical information during reasoning.
We find that language models exhibit progressive attention shifts towards key information during reasoning, which implies essential clues for drawing conclusions.
Building on this observation and analysis, we introduce a novel CoT distillation framework that transfers the teacher's stepwise attention on key information to the student model.
This establishes structured guidance for the student's progressive concentration on key information during reasoning.
More importantly, we develop a Mixture of Layers module enabling dynamic alignment that adapts to different layers between the teacher and student.
Our method achieves consistent performance improvements across multiple mathematical and commonsense reasoning datasets.
To our knowledge, it is the first method to leverage stepwise attention within CoT distillation to improve small model reasoning. 
\end{abstract}

\section{Introduction}
The ability of complex reasoning is a cornerstone of human intelligence, playing a crucial role in problem-solving, decision-making, and world understanding~\cite{cobbe2021training,chu-etal-2024-navigate,plaat2024reasoning}. 
Recent advances have shown substantial improvements in the few-shot reasoning abilities of large language models.
However, the immense scale of these models demands enormous memory and computational resources, making them prohibitively expensive to deploy on edge devices and impeding applications~\cite{liu2024aligning,hu2024minicpm}. 
To address this challenge, CoT distillation~\cite{ho-etal-2023-large, fu2023specializing,li-etal-2023-symbolic,hsieh-etal-2023-distilling} has emerged as a promising approach. 
In complex reasoning, CoT distillation methods typically transfer the step-by-step rationales generated by the teacher model to the student model, serving as an effective means of knowledge distillation.

\begin{figure}[!tb] 
    \centering
    \begin{subfigure}[t]{\linewidth}
    \includegraphics[width=\linewidth]{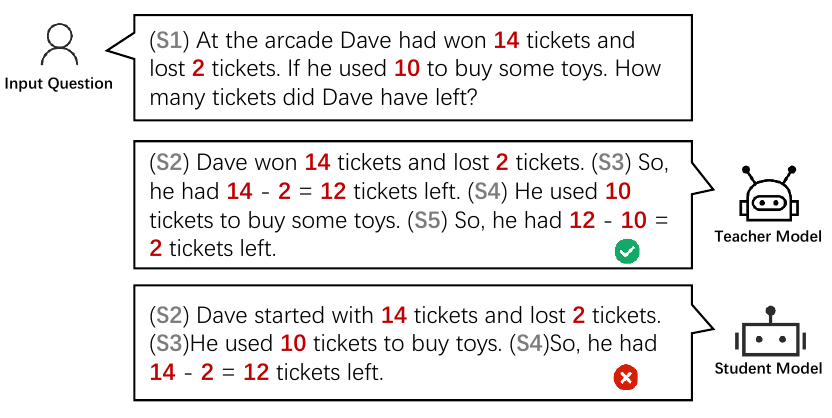} 
        \caption{A sample from the SVAMP dataset. The distilled student model fails to adequately utilize numerical information, leading to erroneous results, whereas the teacher model, during stepwise reasoning, effectively utilizes all numerical information to arrive at the correct final result.} \label{fig:intro1}
    \end{subfigure}
    \vspace{2ex} 
    \begin{subfigure}[t]{\linewidth} 
    \includegraphics[width=\linewidth]{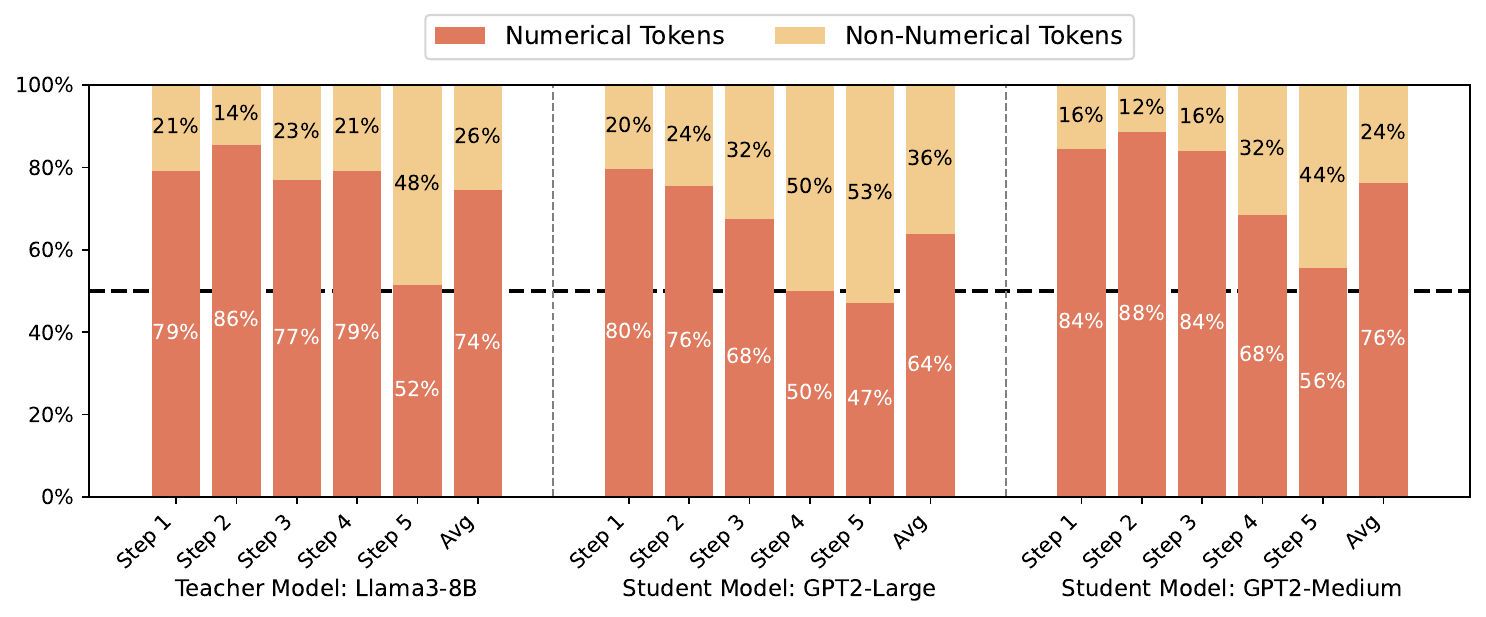} 
        \caption{Numerical vs. Non-Numerical Tokens in Mathematical
        Reasoning: The horizontal axis represents the reasoning steps,
        and the vertical axis shows the relative proportion of stepwise
        attention received by numerical and non-numerical tokens,
        respectively (details in {Appendix~\ref{app: Mathematical Reasoning}}).} \label{fig:intro3}
    \end{subfigure}
    \vspace{2ex} 
    \begin{subfigure}[t]{\linewidth} 
    \includegraphics[width=\linewidth]{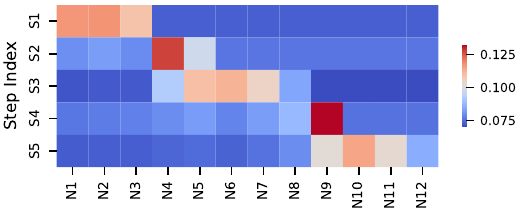} 
        \caption{Visualization of stepwise attention on numerical tokens from the 13th layer of the teacher model Llama3-8B for the sample in Figure {\ref{fig:intro1}} (details in {Appendix~\ref{app: Progressive Attention Pattern on Critical Tokens}}). The horizontal axis represents the indices of numerical tokens (the tokens highlighted in \textbf{\textcolor[HTML]{C72524}{red}} in sample Figure {\ref{fig:intro1}}), and the vertical axis represents the indices of steps (the \textbf{\textcolor[HTML]{A1A1A1}{grey Sx}} labels in Figure {\ref{fig:intro1}}).}  \label{fig:intro2}
    \end{subfigure}
    \vspace{-0.5cm} 
    \caption{Stepwise attention on critical tokens implicitly encodes reasoning clues: A comprehensive analysis.}
\end{figure}

Existing CoT distillation methods typically treat all tokens equally, often neglecting critical information for complex reasoning.
We observe that the student models distilled via existing methods struggle to fully utilize key information across multi-step reasoning (Figure \ref{fig:intro1}).
Notably, language models allocate more average attention to critical tokens during reasoning, implicitly encoding key clues for stepwise reasoning.
For example, numerical tokens are intuitively crucial for mathematical reasoning, and our analysis results indicate that they indeed receive significantly more attention than non-numerical tokens during this process in both teacher and student models (Figure {\ref{fig:intro3}}).
More importantly, we explore how the teacher model’s attention to these critical tokens evolves during stepwise reasoning, and find that the attention distribution exhibits stepwise changes, with higher attention scores assigned to the critical tokens relevant to each reasoning step (Figure {\ref{fig:intro2}} \& Figure {\ref{fig: qwen_layer}}).
This highlights the teacher model's ability to progressively capture key information during reasoning. 
However, current CoT distillation methods directly provide the rationales generated by the teacher model to the student. 
This approach fails to fully exploit the aforementioned phenomena, leading to a failure in improving the student's ability to progressively capture and utilize key information.

Building on the above insights, we introduce \textbf{MoLSAKI}, a novel CoT distillation framework that captures and transfers the teacher model’s \textbf{\underline{S}}tepwise \textbf{\underline{A}}ttention on \textbf{\underline{K}}ey \textbf{\underline{I}}nformation to enhance the student model’s reasoning capabilities via a \textbf{\underline{M}}ixture-\textbf{\underline{o}}f-\textbf{\underline{L}}ayers alignment strategy.
Specifically, we define stepwise attention on critical tokens as the attention weights assigned to each critical token at each reasoning step. 
By concatenating these per‐step distributions, we capture the model’s evolving focus on key information throughout the entire reasoning process. 
Building on this concept, we then extract these stepwise attention maps from every layer of both the teacher and student models during the CoT distillation.
For layer mapping in distillation, we design Mixture-of-Layers (MoL), drawing inspiration from Mixture-of-Experts (MoE)~\cite{zhou2022mixture,jin2024moh}. MoL facilitates adaptive weighted alignment between teacher and student layers, thereby overcoming the distillation challenge of mismatched layer counts.
In summary, our contributions are as follows: 
\begin{itemize}[nosep, leftmargin=*]
\item We introduce a new perspective: during the reasoning process, large language models exhibit a progressive attention pattern towards certain critical tokens, a pattern that implicitly encodes valuable clues for stepwise reasoning.
\item We propose a novel chain-of-thought distillation framework, MoLSAKI, which introduces the concept of stepwise attention on critical tokens and transfers the teacher model’s progressive, dynamic focus on key information to the student model, thereby enhancing its capacity for effective reasoning.
\item We design MoL to adaptively align layers between teacher and student models of different depths in a weighted and dynamic manner, thereby successfully overcoming the challenge of their mismatched layer counts.
\item Our method yields performance gains in in-domain and out-of-domain settings across varying teacher-student model scales on mathematical and commonsense reasoning benchmarks.
\end{itemize}

\section{Related Work}
\subsection{Chain-of-Thought Distillation}
Large language models (LLMs) demonstrate strong reasoning capabilities~\cite{kojima2022large,wei2022chain}, yet their massive scale hinders practical deployment. Recent work distills reasoning abilities into smaller models through CoT knowledge transfer~\cite{ho-etal-2023-large,hsieh-etal-2023-distilling,fu2023specializing,li-etal-2023-symbolic}. Key approaches include Fine-tune-CoT's zero-shot rationale extraction~\cite{ho-etal-2023-large} and DSS's multi-task separation of reasoning/answer prediction~\cite{hsieh-etal-2023-distilling}. Subsequent improvements introduce mutual information maximization (MMIloss~\cite{chen-etal-2024-learning-maximize}) and auxiliary model-based distillation (Mentor-KD~\cite{lee2024mentor}) (details in {Appendix~\ref{subapp: Chain of Thought Distillation}}).
Existing methods neglect key information in reasoning and face structural constraints from logit distillation requirements~\cite{lee2024mentor,zhang-etal-2024-dual}. Our approach introduces stepwise attention on critical tokens distillation without requiring tokenizer alignment or projection layers.

\begin{figure}[!tb] 
    \centering
    \begin{subfigure}[t]{\linewidth}
    \includegraphics[width=\linewidth]{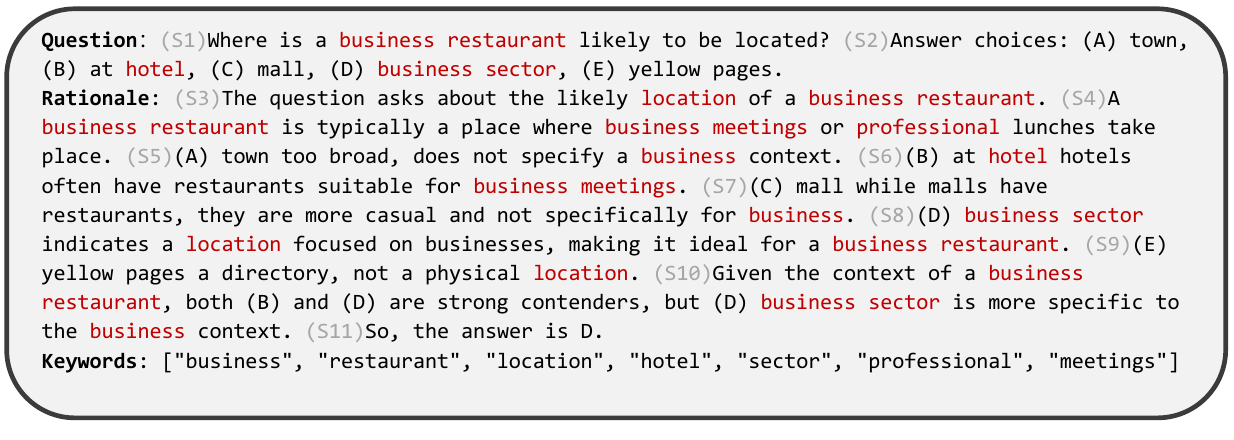} 
        \caption{A sample from the CommonSenseQA dataset.} \label{fig: qwen_sample}
    \end{subfigure}
    \vspace{2ex} 
    \begin{subfigure}[t]{\linewidth} 
    \includegraphics[width=\linewidth]{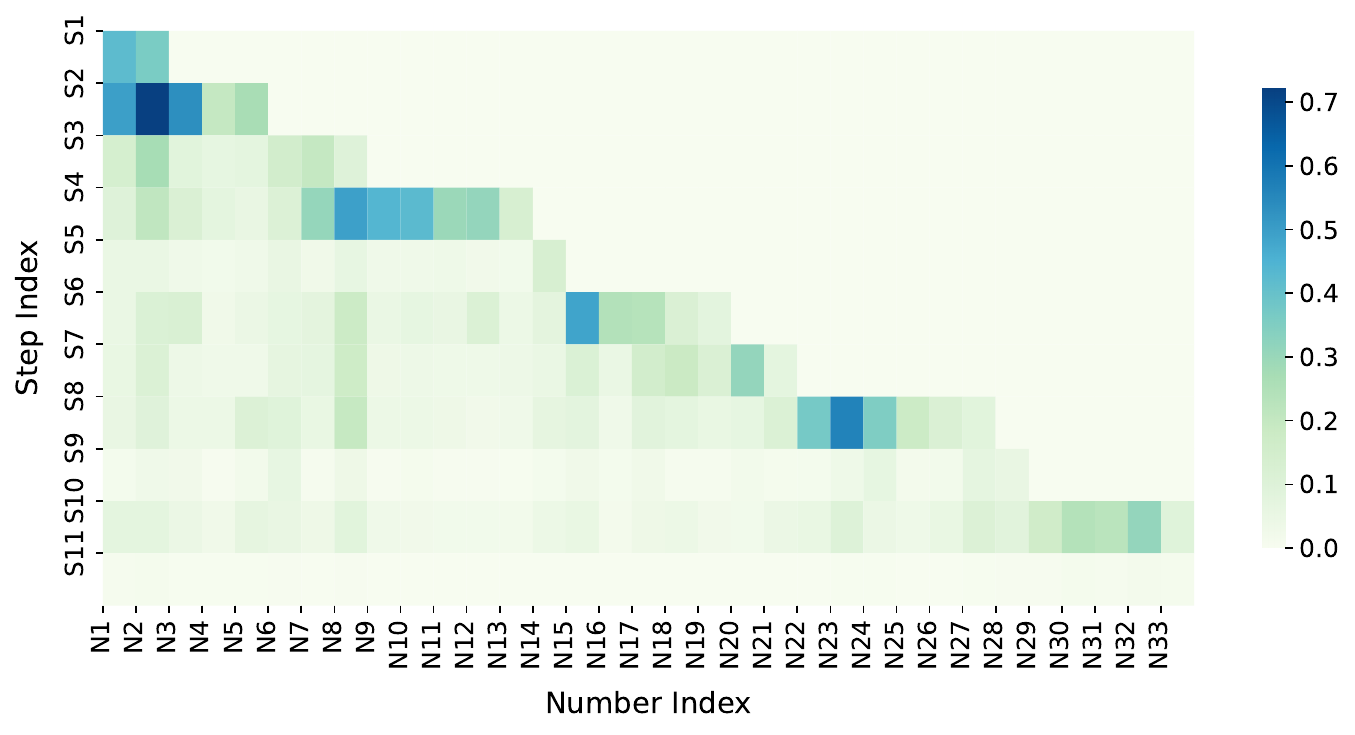} 
        \caption{Visualization of stepwise attention on critical tokens
from the 32nd layer of the teacher model Qwen2.5-32B for the
sample in Figure \ref{fig: qwen_sample}. The horizontal axis represents the indices of critical tokens, and the vertical axis represents the indices of steps.} \label{fig: qwen2}
    \end{subfigure}
    \vspace{-0.5cm} 
    \caption{Progressive attention pattern on critical tokens (details in Appendix~\ref{app: Progressive Attention Pattern on Critical Tokens}).}
    \label{fig: qwen_layer}
\end{figure}

\subsection{Self-Attention Distillation}
Prior methods transfer self-attention patterns via layer mapping: TinyBERT~\cite{jiao2020tinybert} uses uniform mapping, MOBILEBERT~\cite{sun2020mobilebert} assumes identical layer counts, and MINILM~\cite{wang2020minilm} distills only final layers (details in {Appendix~\ref{subapp: Self-Attention Distillation}}). These methods require matched attention dimensions and fixed layer correspondences.
We overcome these limitations by 1) focusing distillation on critical tokens in reasoning steps instead of full attention matrices, and 2) using dynamic layer routing via MoL modules to automatically select optimal teacher-student layer pairs, outperforming rigid mapping approaches.

\begin{figure*}[!htb]
  \includegraphics[width=\linewidth]{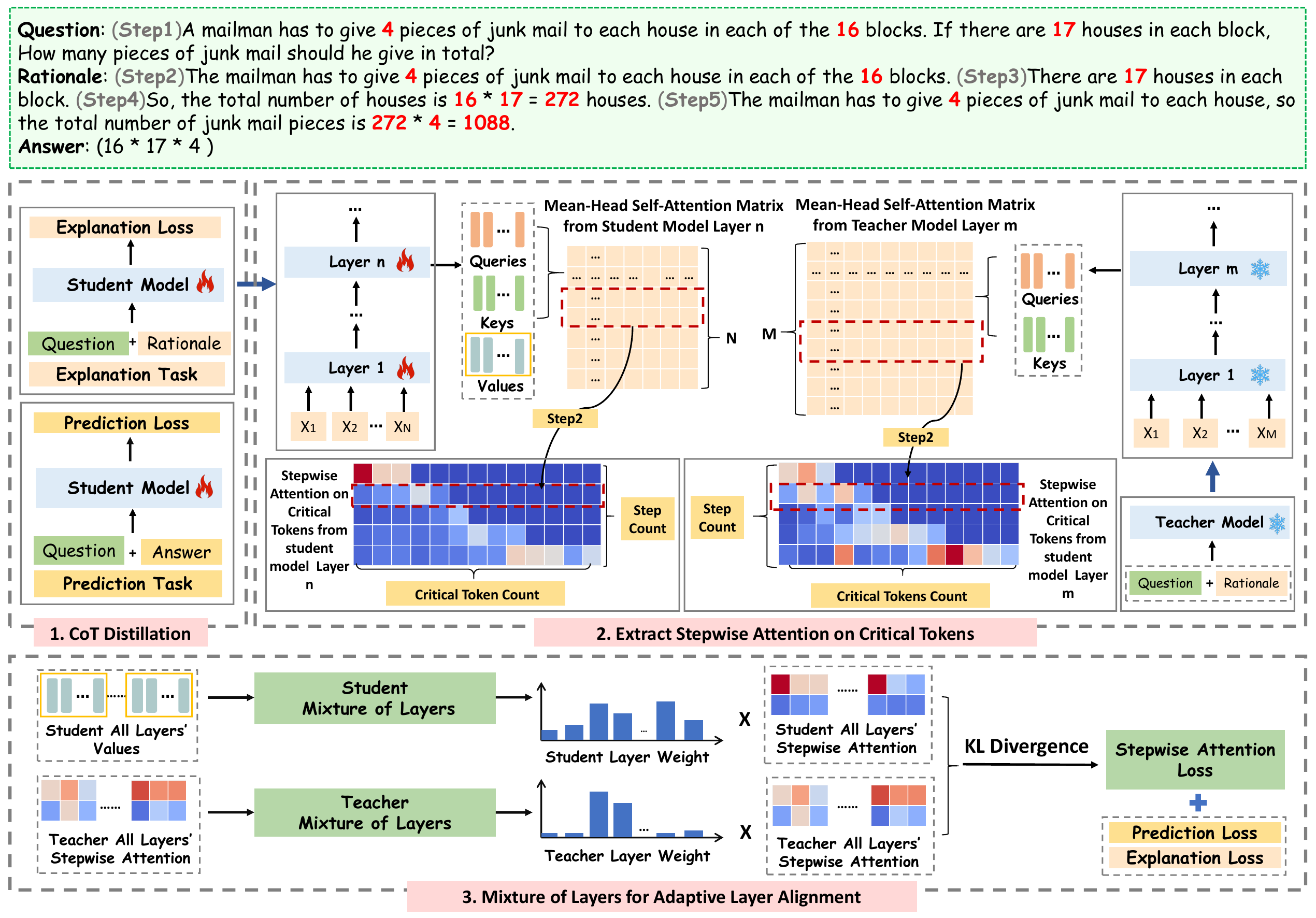} 
  \caption {The MoLSAKI framework consists of three components. In the example, the question and rationale have \underline{13} numerical tokens and \underline{5} steps in total. Thus, the stepwise attention on numerical tokens in both teacher and student models is \underline{5}$\times$\underline{13}.}
  \label{Framework}
\end{figure*}

\section{Methodology}
MoLSAKI introduces a novel knowledge distillation framework that enhances the reasoning of the student model through synergistic integration of CoT distillation and stepwise attention guidance.
Specifically, we first prepare CoT data annotated by the teacher model and conduct CoT distillation ({\S}\textbf{\ref{sub3.1}}), subsequently extract stepwise attention on critical tokens from the teacher and student models in the process of CoT distillation ({\S}\textbf{\ref{sub3.2}}), and finally implement adaptive MoL layer alignment ({\S}\textbf{\ref{sub3.3}}).

\subsection{CoT Distillation}
\label{sub3.1}
We obtain CoT data for each question-answer pair $\{q, \hat{a} \}$ in a raw dataset $\mathcal{D}$ by few-shot prompting the teacher model (details in {Appendix~\ref{app: Prompts}}).
The teacher's response to each question $q$ is divided into two components: rationale $r$ and answer $a$ (see the sample in Figure \ref{Framework}).
The labeled dataset $\{q,r,a ~|~q\in\mathcal{D},~a=\hat{a}\}$ will be used for the subsequent CoT distillation of the student model.

Following \citet{hsieh-etal-2023-distilling}, we perform CoT distillation comprising two tasks (\textit{CoT Distillation} module in Figure {\ref{Framework}}): 1) final answer prediction $a$ given a question $q$ and 2) rationale $r$ generation for the same input  $q$.
The respective loss functions are as follows:
\begin{equation} \label{Eq:loss} \small
    \begin{aligned}
\mathcal{L}_{\text{pre}} = \mathbb {E}_{q \in \mathcal{D}}\;[\mathcal{L}_{\text{ce}}(f(q), {a})], \\
\mathcal{L}_{\text{exp}} = \mathbb {E}_{q \in \mathcal{D}}\;[\mathcal{L}_{ce}(f(q), {r})],
    \end{aligned}
\end{equation}
where $f$ denotes the student model and $\mathcal{L}_{\text{ce}}$ denotes the cross-entropy loss between model predictions and target tokens.

\subsection{Stepwise Attention on Critical Tokens}
\label{sub3.2}
Believing that distilling the teacher's stepwise attention on critical tokens during reasoning is more impactful than simply transferring rationales, we introduce the loss $L_{att}$ (in Eq.(\ref{eq:latt})) of stepwise attention on critical tokens during CoT distillation to guide the student's progressive focus on key information.

To compute the loss $L_{att}$, we first extract stepwise attention on critical tokens from both the teacher and student models (\textit{Extract Stepwise Attention on Critical Tokens} module in Figure {\ref{Framework}}). 
In our design, \textit{Stepwise} denotes reasoning steps incorporating the question. As shown in the example in Figure \ref{Framework}, we segment the input sequence composed of question and rationale into reasoning steps based on periods, resulting in 5 steps.

The teacher model's tokenizer converts the input sequence composed of question and rationale into a token sequence $\{x_1^t, x_2^t,...,x_M^t\}$.
$\mathcal{M}_1$ denotes the index set of all tokens partitioned by reasoning steps. Its element specifically denotes the index set of all tokens within a single reasoning step. Utilizing regular expression matching and the tokenizer's mapping, we obtain the index set of critical tokens from the token sequence, denoted as $\mathcal{M}_2$. Its element denotes the index set of critical tokens corresponding to a specific critical word in the original text after tokenization (details in {Appendix~\ref{app: Extract Indices of Critical Tokens}}).

The $l$-th layer of the teacher model subsequently constructs the self-attention matrix $I_{l}^{t}\in\mathbb{R}^{M\times M}$. To compute stepwise attention on critical tokens, we first extract columns from $I_{l}^{t}$ at the indices of critical tokens, where each column represents attention distribution from all tokens to a specific critical token.
Based on this, we compute the aggregated stepwise attention on critical tokens by summing rows of the corresponding columns in  $I_{l}^{t}$ at each reasoning step, as follows:
\begin{equation}\small
\begin{aligned}
&\mathcal{M}_1:= \{\{0,1,2,...\},...,\{...,M-1\} \},\\
&\mathcal{M}_2:= \{\{\mu_1,\mu_2,...\},...\},\ \mu\in \mathbb{N} ,\ \mu<{M},\\
&{A}^t_{l}=\sum_{i\in\mathcal{K},\ j\in\mathcal{P}} I^{t}_{l}[i, j]\ (  \mathcal{K}\in \mathcal{M}_1, \mathcal{P} \in \mathcal{M}_2),
\end{aligned}
\label{Eq:cal_sa_t}
\end{equation}
where ${A}^t_{l}\in \mathbb{R}^{|\mathcal{M}_1|\times|\mathcal{M}_2|}$ corresponds to the stepwise attention scores on critical tokens generated by the $l$-th layer of the teacher model.

The student model processes the token sequence  $\{x_1^s, x_2^s,...,x_N^s\}$ to generate self-attention matrix $I_l^s \in \mathbb{R}^{N\times N}$ of the $l$-th layer, from which we apply the identical extraction and aggregation mechanism to compute its stepwise attention on critical tokens:
\begin{equation}\small
\begin{aligned}
&\mathcal{N}_1:= \{\{0,1,2,...\},...,\{...,N-1\} \},\\
&\mathcal{N}_2:= \{\{\lambda_1,\lambda_2,...\},...\},\ \lambda\in \mathbb{N} ,\ \lambda<{N},\\
&{A}^s_{l}=\sum_{i\in\mathcal{K}\,j\in\mathcal{O}} I^{s}_{l}[i, j]\ (\mathcal{K}\in\mathcal{N}_1,\ \mathcal{O} \in \mathcal{N}_2),
\label{Eq:cal_sa_s}
\end{aligned}
\end{equation}
where ${A}^s_{l}\in \mathbb{R}^{|\mathcal{N}_1|\times|\mathcal{N}_2|}$ denotes the stepwise attention scores on critical tokens generated by the $l$-th layer of the student model with $|\mathcal{N}_1|=|\mathcal{M}_1|$  indicating the total count of reasoning steps and $|\mathcal{N}_2|=|\mathcal{M}_2|$ representing the total count of critical tokens (details in {Appendix~\ref{app: Identical Shape}}). Our mechanism achieves functional compatibility between architecturally distinct models by aligning the stepwise attention dimensions on critical tokens across teacher and student models, thereby eliminating the requirement for shared tokenizers or vocabularies.

\subsection{MoL for Adaptive Layer Alignment}
\label{sub3.3}
Though distilling stepwise attention on critical tokens is feasible, determining optimal layer mapping in distillation presents a non-trivial challenge. 
This arises from architectural disparities between teacher and student models, which preclude complete layer-to-layer correspondence. Conventional rigid Single-Layer (SL) alignment approaches prove suboptimal due to their inflexibility. 
To overcome these limitations, we propose a Mixture-of-Layers (MoL) module that dynamically aggregates stepwise attention across all layers through trainable weighting parameters in the layer router.

\begin{figure}[!htb]\centering
\includegraphics[width=0.92\linewidth]
{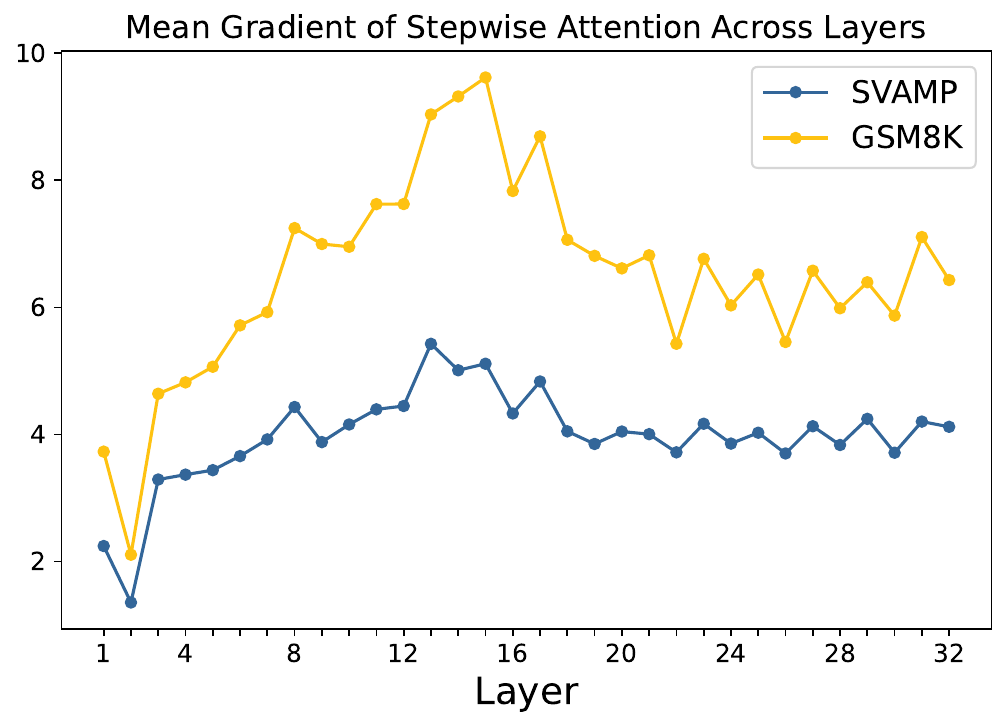} 
  \caption {We analyse the average column gradient distribution of stepwise attention on critical tokens across the layers of Llama3-8B (details in {Appendix~\ref{app: Teacher}}).}
  \label{fig:teacher_gradient}
\end{figure}

Leveraging insights from the analysis of teacher and student model characteristics, distinct inputs are provided to the teacher and student MoL modules (\textit{Mixture of Layers for Adaptive Layer Alignment} module in Figure {\ref{Framework}}). For the teacher model, the stepwise attention on critical tokens $A^t_l$ varies across its different layers (Figure {\ref{teacher_layer_attention}}).
By analysing the column gradients of $A^t_l$, we find that the most significant variation of stepwise attention on critical tokens occurs in the intermediate layers (Figure {\ref{fig:teacher_gradient}}).

To effectively transfer the significant attention dynamics to the student model, we determine the teacher model's layer weights through temperature-controlled softmax normalisation applied to the gradients of $A^t_l$, as follows:
\begin{equation}
\small
    \begin{aligned}
    &\mathrm{G}({A}_{l}^{t}) = \frac{\sum_{i=1}^{|\mathcal{M}_1|} \sum_{j=1}^{|\mathcal{M}_2|-1} \left| {A}_{l}^{t}[i,j+1] - {A}_{l}^{t}[i,j] \right|}{|\mathcal{M}_1|(|\mathcal{M}_2|-1)},\label{eq:gradients}\\
    &{p}^{t} = \mathrm{softmax}
    ([\mathrm{G}({A}_{1}^{t}),..,\mathrm{G}({A}_{L_1}^{t})], \tau_1) \in \mathbb{R}^{L_1}, \\
    &\mathrm{softmax}(z, \tau)_i = \frac{e^{z_i/\tau}}{\sum_{j=1}^K e^{z_j/\tau}},
\end{aligned}
\end{equation}
where $\mathrm{G}({A}_{l}^{t})$ denotes the mean gradient of ${A}_{l}^{t}$, $L_1$ indicates the count of layers in the teacher model, the $p^t \in \mathbb{R}^{L_1}$ denotes the layer weights obtained by the MoL of the teacher model, and $\tau_1$ denotes the temperature parameter of the softmax function.

For the student model, we process value vectors from all layers through a learnable routing mechanism:
First, we apply RMSNorm~\cite{zhang2019root} to stabilise the features.
Second, we sum over the sequence dimensions to obtain compact layer embeddings.
Third, we concatenate multi-layer representations.
Finally, we generate adaptive layer weights via an affine transformation and a temperature-controlled softmax.
The above procedure is formulated as:
\begin{equation}\small
    \begin{aligned}
    &\widetilde{V}_l =\mathrm{RMSNorm}(V_l) \in \mathbb{R}^{N\times d}, \\
    &h_l = \sum_{i=1}^{N}(\widetilde{V}_l[i,:]) \in \mathbb{R}^{d}, \\
    &H = \mathrm{concat} (h_1, h_2, ..., h_{L_2}) \in \mathbb{R}^{L_2\times d},  \\
    &p^{s} = \mathrm{softmax}
    (HW + b, \tau_2) \in \mathbb{R}^{L_2},
\end{aligned}
\end{equation}
where $d$ denotes the dimension of the value vector, $L_2$ indicates the count of layers in the student model,  the $p^s\in\mathbb{R}^{L_2}$ denotes the layer weights obtained by the MoL of the student model, and $\tau_2$ denotes the temperature parameter of the softmax function.
Building upon this foundation, we independently applied weighting to  $A^t_l$ and $A^s_l$ respectively, followed by performing softmax normalisation along the temporal step dimension.
Subsequently, the averaged Kullback-Leibler (KL) divergence across corresponding steps is calculated and designated as stepwise attention loss $\mathcal{L}_{att}$: 
\begin{equation}\small
\begin{aligned}
     &{A}^{t} =  \sum_{l=1}^{L_1} (p^{t}_l {A}^{t}_l)\in \mathbb{R}^{|\mathcal{N}_1| \times |\mathcal{N}_2|}, \\
    &\widetilde{A}^t[i,:] = \mathrm{softmax}({A}^t[i,:]), \\
    &{A}^{s} =  \sum_{l=1}^{L_2} (p^{s}_l {A}^{s}_l)\in \mathbb{R}^{|\mathcal{N}_1| \times |\mathcal{N}_2|}, \\
    &\widetilde{A}^s[i,:] = \mathrm{softmax}({A}^s[i,:]),\\
    &\mathcal{L}_{att} = \frac{1}{|\mathcal{N}_1|}\sum_{i=1}^{|\mathcal{N}_1|} \mathrm{KL}(\widetilde{A}^{t}[i,:] \parallel \widetilde{A}^{s}[I,:]).
    \label{eq:latt}
\end{aligned}
\end{equation}

Finally, we formulate the overall objective function $\mathcal{L}$ through a weighted combination as:
\begin{equation}\small
    \mathcal{L} = \alpha \mathcal{L}_{\text{pre}} + (1-\alpha) \mathcal{L}_{\text{exp}} + \beta \mathcal{L}_{\text{att}}.
    \label{Eq:loss_all}
\end{equation}
where the prediction loss $\mathcal{L}_{\text{pre}}$ and the explanation loss $\mathcal{L}_{\text{exp}}$ are in Eq.(\ref{Eq:loss}), and $\mathcal{L}_{\text{att}}$ is the aforementioned stepwise attention loss.

\section{Experiments}
\begin{table}
  \centering
    \footnotesize
    \setlength{\tabcolsep}{1.1pt} 
  \begin{tabular}{l|ccccc}
    \toprule
    \midrule
      &\textbf{SVAMP}  &\textbf{SingleEq}  &\textbf{AsDiv} &\textbf{GSM8K} &\textbf{CSQA} \\
    \midrule
    \textbf{In-Domain}   &\ding{51}  &\ding{55}  &\ding{55}  &\ding{55}  &\ding{51}   \\
     \midrule
      \multicolumn{6}{c}{\textbf{Teacher:Llama3-8B Student:GPT2-Large}} \\
     \midrule
    Vanilla Finetune &10 &12.1 &9.2 &4.2  & 16.7  \\
    DSS &48.0  &36.1 &30.3  &12.4 &19.1   \\
    MMIloss &47.0  &37.9 &30.7 &12.5  &19.4   \\
    \rowcolor{gray!20}
    MoLSAKI(ours) &\textbf{49.5} &\textbf{39.8} &\textbf{32.2} &\textbf{15.1} &\textbf{21.0 }   \\
    \midrule
      \multicolumn{6}{c}{\textbf{Teacher:Qwen2.5-32B Student:TinyLllama-1.1B}} \\
     \midrule
    Vanilla Finetune  &14.5 &21.4 &14.3 &6.7 &17.8  \\
    DSS &59.5 &48.1 &33.5  &13.8 &28.9	  \\
    MMIloss &64.5	&48.1	&42.6 &14.0  &25.8  \\
    \rowcolor{gray!20}
    MoLSAKI(ours)&\textbf{68.5}&\textbf{51.8} &\textbf{43.3} &\textbf{16.9}  &\textbf{30.3} \\
    \midrule
    \bottomrule
  \end{tabular}
  \caption{Accuracy(\%) of different approaches.}
  \label{main_result}
\end{table}

\subsection{Setup}
\noindent{\textbf{Datasets. }} In the experiment, five public reasoning datasets are utilized: SVAMP~\cite{patel-etal-2021-nlp}, SingleEq~\cite{koncel-kedziorski-etal-2015-parsing}, Asdiv~\cite{miao2021diverse}, GSM8K~\cite{cobbe2021training}, CommonSenseQA (CSQA)~\cite {talmor-etal-2019-commonsenseqa}.
To assess the effect of our method on the generalization of the student model, we established an in-domain and out-of-domain evaluation setting using mathematical reasoning datasets. For mathematical reasoning, SVAMP is used as an in-domain test dataset, and SingleEq, Asdiv, and GSM8K serve as out-of-domain test datasets. For commonsense reasoning, CSQA is used as an in-domain test dataset. (details in Appendix~\ref{Dataset Statistics}).

\noindent{\textbf{Baselines. }} We compare our proposed MoLSAKI framework with three established baseline methods: 1) \textit{Vanilla Fine-Tuning}  ($\mathcal{L}=\mathcal{L}_\text{pre}$) trains models exclusively on answer labels without CoT utilization;  2) \textit{DSS}~\cite{hsieh-etal-2023-distilling} ($\mathcal{L} = \alpha \mathcal{L}_{\text{pre}} + (1-\alpha) \mathcal{L}_{\text{exp}}$) conducts multi-task distillation that decouples rationale and answer optimisation; 3) \textit{MMIloss}~\cite{chen-etal-2024-learning-maximize} ($\mathcal{L} = \alpha \mathcal{L}_{\text{pre}} + (1-\alpha) \mathcal{L}_{\text{exp}} + \beta \mathcal{L}_{\text{MMI}}$) extends DSS by incorporating cross-entropy loss between rationale generation and answer prediction as an auxiliary objective under the information bottleneck principle. In the experiments, we followed the default hyperparameter settings of these works, using $\alpha=0.5,\beta=0.1$.

\noindent{\textbf{Settings. }} In the main experiments, we employed two teacher-student model configurations: (1) Llama3-8B~\cite{Meta2024} as the teacher model and GPT-2 Large (774M)~\cite{radford2019language} as the student model; and (2) Qwen2.5-32B~\cite{qwen2025qwen25technicalreport} as the teacher model and TinyLlama-1.1B~\cite{zhang2024tinyllama} as the student model. 
For the analysis experiment, the teacher model was Llama3-8B, and the student model was GPT-2 Medium (355M). 
For the main experiments, the weight hyperparameter $\beta$  of the stepwise attention loss was set to $1.0$, and the temperature hyperparameters for the MoL of the teacher and student models were set to $\tau_1=0.1$ and $\tau_2=0.5$, respectively (details in {Appendix~\ref{app: Temperature Parameters}}).


\subsection{Main Results}
In this section, we thoroughly evaluate MoLSAKI through in-domain and out-of-domain tests. The results demonstrate that it works effectively and maintains consistent performance across different reasoning datasets, demonstrating its reliability.

1) \textit{MoLSAKI substantially boosts student models' reasoning performance.} 
CoT distillation methods notably improve the performance of student models in reasoning tasks compared to the standard fine-tuning approach, as shown in Table~\ref{main_result}. 
Our proposed MoLSAKI method achieves an average relative improvement of 7.5\% (GPT2-Large) and 11.3\% (TinyLlama) over the baselines. A detailed computational comparison of the methods and further case studies is provided in {Appendix~\ref{app: Computational Cost}} and {Appendix~\ref{app: Case Study}}.

2) \textit{MoLSAKI consistently achieves significant in-domain accuracy improvements across varying model scales.} Specifically, on the in-domain datasets, it outperforms both the DSS and MMIloss baselines for two distinct student model scales. This superior performance is attributed to MoLSAKI's ability to improve knowledge transfer by guiding the student's attention at each reasoning step through stepwise attention alignment.

3) \textit{MoLSAKI demonstrates strong generalization capabilities on out-of-domain reasoning tasks.} Experimental results consistently show that our method outperforms baseline approaches across out-of-domain benchmarks for two different student models. This superior out-of-domain generalization underscores the importance of distilling the teacher model's stepwise attention focusing on critical tokens.

\subsection{Hyperparameter Analysis}

\begin{figure}[!htb] 
    \centering
    \begin{minipage}[b]{\linewidth}
        \centering
        \includegraphics[width=\linewidth]{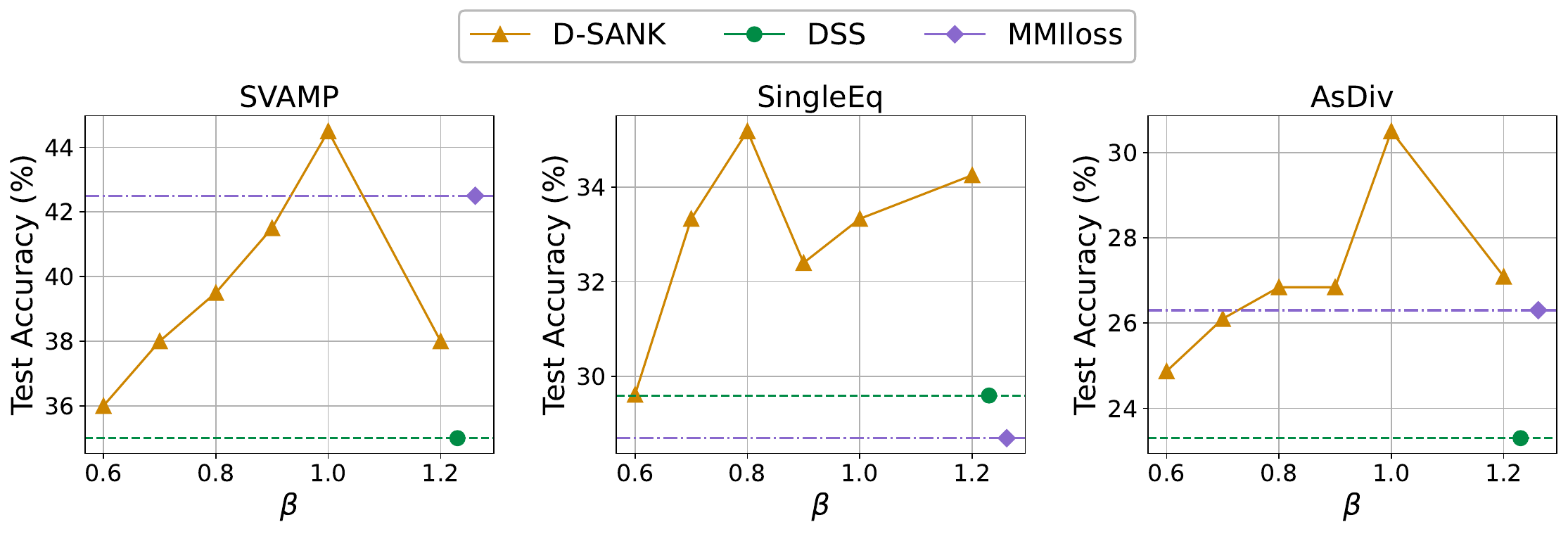} 
    \end{minipage}
    
    \vspace{0cm}  
    \begin{minipage}[b]{\linewidth} 
        \centering
        \includegraphics[width=\linewidth]{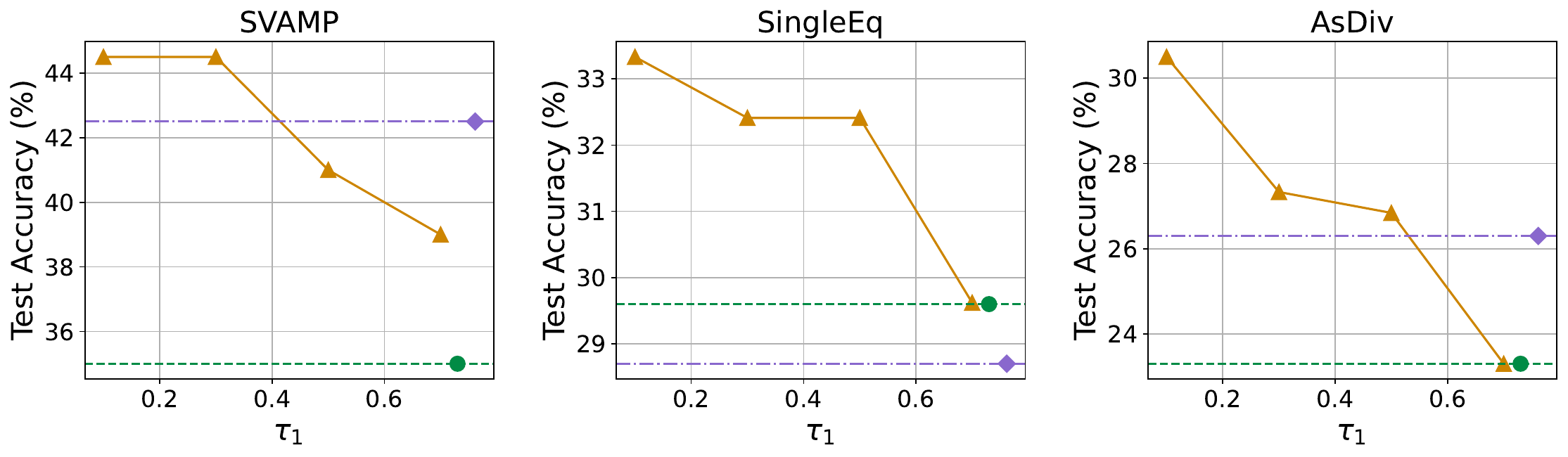} 
    \end{minipage}
    
    \vspace{0cm}  
    \begin{minipage}[b]{\linewidth}
        \centering
        \includegraphics[width=\linewidth]{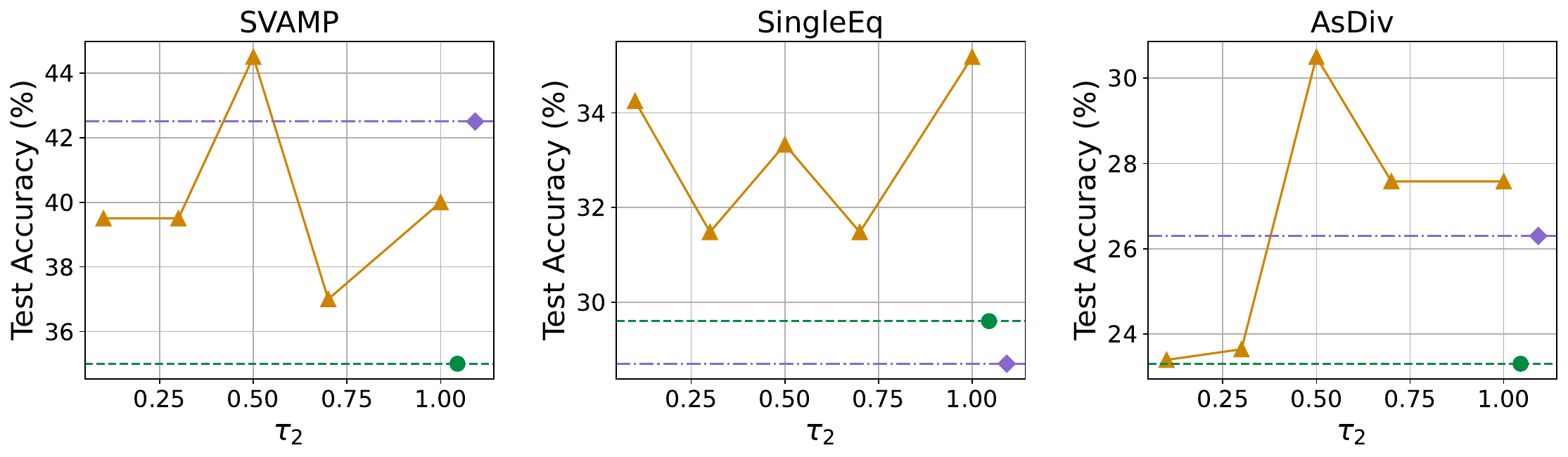} 
    \end{minipage}
    
    \caption{Hyperparameter analysis of ${\beta}$, ${\tau_1}$, and ${\tau_2}$ in MoLSAKI.}
    \label{hyperparameter_analysis}
\end{figure}

In this section, we systematically assess the impact of the weight hyperparameter $\beta$ of stepwise attention loss $\mathcal{L}_{\text{att}}$, as well as the temperature parameters $\tau_1$ and $\tau_2$ modulating the layer weight of the teacher and student models respectively. 
We conducted experiments using the student model GPT2-Medium and the teacher model Llama3-8B on three mathematical reasoning datasets, where SVAMP serves as the in-domain test dataset and SingleEq and AsDiv are out-of-domain test datasets.

Our $\beta$ analysis reveals that MoLSAKI's SVAMP performance peaks when $\beta$ is set to 1.0 before declining (Figure~\ref{hyperparameter_analysis}), surpassing MMIloss only at this optimal value. 
However, it maintains consistent advantages across out-of-domain datasets under most $\beta$ settings, further demonstrating that MoLSAKI enhances the generalisation of the student model's mathematical reasoning ability.

For $\tau_1$ of MoL in the teacher model, the performance of MoLSAKI gradually declines as $\tau_1$ increases (Figure~\ref{hyperparameter_analysis}). 
This suggests that during the stepwise attention distillation process, the teacher model tends to prioritise layers with larger stepwise attention gradients, enabling these layers to contribute more prominently to the transfer of attention information in the distillation process. 
For $\tau_2$ of MoL in the student model, the results on the SVAMP and AsDiv datasets reach their maximum values when $\tau_2$ is set to 0.5 (Figure~\ref{hyperparameter_analysis}). 
Extreme $\tau_2$ values degrade MoLSAKI performance, indicating that stepwise attention distillation necessitates balanced layer participation in the student model while preventing excessive uniformity.
Temperature hyperparameters exhibit patterns comparable to $\beta$: while suboptimal configurations cause SVAMP performance to dip below MMIloss, most settings achieved superior generalisation.

\subsection{Layer Weight Visualization}
\label{sec: Layer Weight Visualization}
\begin{figure}[!htb] 
    \centering
    \begin{minipage}[b]{\linewidth}
        \centering
        \includegraphics[width=\linewidth]{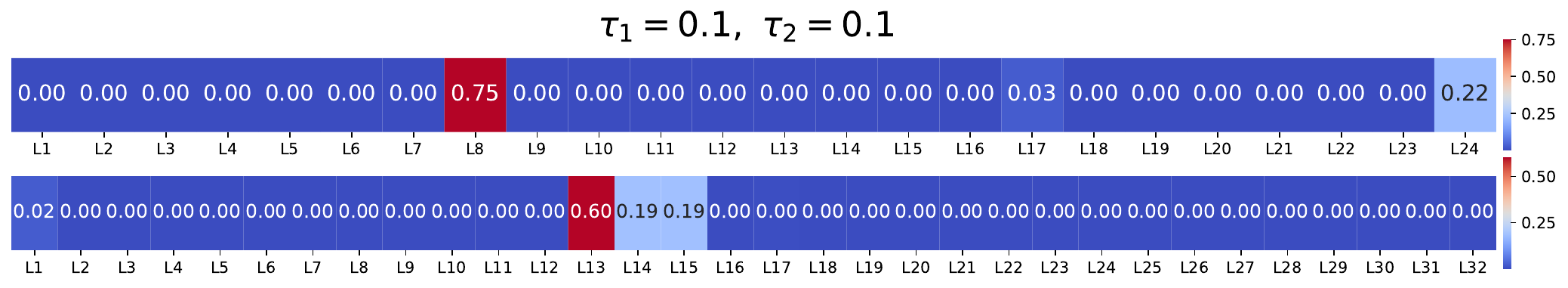} 
    \end{minipage}
    
    \vspace{0cm}  
    \begin{minipage}[b]{\linewidth} 
        \centering
        \includegraphics[width=\linewidth]{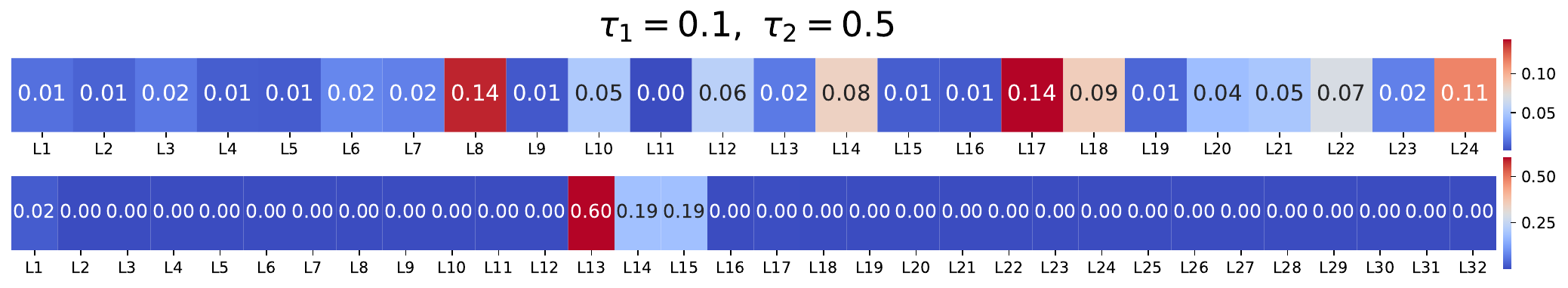}
    \end{minipage}
    
    \vspace{0cm}  
    \begin{minipage}[b]{\linewidth} 
        \centering
        \includegraphics[width=\linewidth]{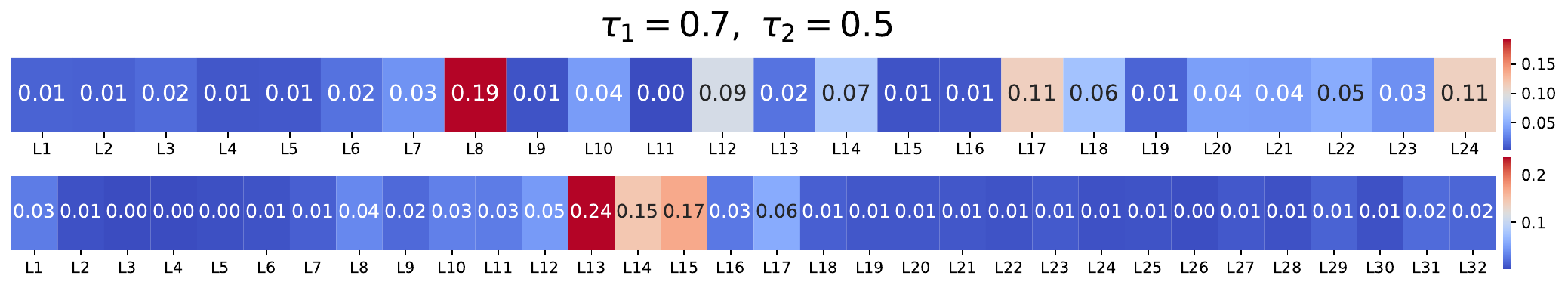}
    \end{minipage}
    
    \caption{Comparative visualization of layer weight in Llama3-8B (32-layer) and GPT2-Medium (24-layer) under parameter configurations $\tau_1$ and $\tau_2$.}
    \label{visual_weight}
\end{figure}
This section visualises layer weight distributions in MoLSAKI's MoL under three $(\tau_1,\tau_2)$ configurations, comparing teacher (Llama3-8B) and student (GPT2-Medium) models in Figure~\ref{visual_weight}.

When setting temperature coefficients to $\tau_1=0.1$ and $\tau_2=0.5$, the teacher model exhibits significant layer-weight differentiation during stepwise attention distillation (details in {Appendix~\ref{app: Temperature Parameters}}). 
Specifically, layers 13-15 demonstrate maximum weight values, while other layers show parameter attenuation approaching zero. 
In contrast, the student model maintains relatively balanced weight distribution throughout the distillation process: although layers 8, 17, and 24 attain comparatively higher weights, the remaining layers preserve non-negligible values, forming a distinct contrast with the near-zero weight pattern observed in most teacher model layers.

With temperature parameters  $\tau_1=0.1$ and $\tau_2=0.1$, the student model demonstrates pronounced weight concentration during distillation: layer 8 emerges as the dominant contributor with maximum weight magnitude, followed distantly by layer 24, while all other layers' weights approach negligible values.
In contrast, when configuring $\tau_1=0.7$ and $\tau_2=0.5$, the teacher model exhibits fundamental shifts in weight dynamics: the previously dominant layers 13-15 lose their absolute predominance, giving way to more balanced inter-layer weight allocation.
Additionally, we visualise the layer weight for the teacher and student models under $\tau _1=0.1$ and $\tau _2=1.0$, with the results presented in {Appendix~\ref{Layer Weight Visualization}}.

\subsection{SL Alignment vs. MoL Alignment}
\label{Sec: Fixed Layer Alignment vs Adaptive Layer Alignment}

\begin{table}
  \centering
  \small
  \setlength{\tabcolsep}{3pt}
  \begin{tabular}{l|c|ccc}
    \toprule
    Method & T $\rightarrow$ S & SVAMP & AsDiv & AVG \\ 
    \midrule
    DSS & - & 35.0 & 23.3 & 29.1  \\ 
    \midrule
    MMIloss & - & 42.5 & 26.3 & 34.4 \\ 
    \midrule
    \multirow[c]{11}{*}{\raisebox{-8.2\height}{MoLSAKI}} 
        & 13 $\rightarrow$ 8 & 36.5 & 25.1 & 30.8 \\ 
    \cmidrule{2-5}
        & 13 $\rightarrow$ 17 & 38.5 & 26.6 & 32.5 \\ 
    \cmidrule{2-5}
        & 13 $\rightarrow$ 24 & 35.0 & 26.1 & 30.5 \\ 
    \cmidrule{2-5}
        & 14 $\rightarrow$ 8 & 36.0 & 26.6 & 31.3 \\ 
    \cmidrule{2-5}
        & 14 $\rightarrow$ 17 & 42.5 &\cellcolor{cyan!5}\underline{28.3} & 35.4 \\ 
    \cmidrule{2-5}
        & 14 $\rightarrow$ 24 & 43.0 & 28.1 & 35.5 \\ 
    \cmidrule{2-5}
        & 15 $\rightarrow$ 8 & 35.5 & 27.3 & 31.4 \\ 
    \cmidrule{2-5}
        & 15 $\rightarrow$ 17 & 36.5 & 26.6 & 31.5 \\ 
    \cmidrule{2-5}
        & 15 $\rightarrow$ 24 & 39.0 & 24.6 & 31.8 \\ 
    \cmidrule{2-5}
        & 32 $\rightarrow$ 24 & \cellcolor{cyan!5}\underline{43.0} & 28.1 & \cellcolor{cyan!5}\underline{35.8} \\ 
    \cmidrule{2-5}
        & MoL & \cellcolor{green!5}\textbf{44.5} & \cellcolor{green!5}\textbf{30.5} &\cellcolor{green!5}\textbf{37.5} \\ 
    \bottomrule  
  \end{tabular}
  \caption{Ablation comparison experiment results of adaptive weighted layer alignment (MoL) and fixed single-layer mapping (SL). Boldface denotes the best performance, while underlining denotes the second best.} 
  \label{layer_map}
\end{table}

To validate the effectiveness of the MoL module, we implement two fixed single-layer (SL) alignment strategies for comparison with our adaptive weighted layer alignment (MoL) method:
1) Based on the layer visualization results with $\tau_1=0.1,~\tau_2 =0.5$ (in Sec.\ref{sec: Layer Weight Visualization}), we select the top three highest-weighted layers from both the teacher model Llama3-8B (layers 13, 14, 15) and the student model GPT2-Medium (layer 8, 17, 24).
2) Following MINILM~\cite{wang2020minilm}, we separately select the last layer of the teacher model Llama3-8B (layers 32) and the student model GPT2-Medium (layer 24).

The experimental results demonstrate that MoL's adaptive weighted layer alignment mechanism outperforms conventional fixed single-layer alignment approaches (Table~\ref{layer_map}). Notably, even with the simplified single-layer alignment configuration, some of it still surpasses baseline methods.
This finding suggests that even if only a specific layer of the student model learns to capture the teacher model's specific layer's attention on critical tokens at the step level, it still contributes to the student model’s final reasoning.

\subsection{Rationale and Stepwise Attention Derives from  Different Models}
\begin{table}
  \small
  \centering
    \renewcommand{\arraystretch}{1.12}  
  \setlength{\tabcolsep}{3pt}
  \begin{tabular}{l|cccc}
    \toprule
    Method &SVAMP  &SingleEq  &AsDiv  &AVG \\
    \midrule
     In-Domain   &\ding{51}  &\ding{55}  &\ding{55} &-\\
    \midrule
      \multicolumn{5}{c}{\textbf{Rationale} $\boldsymbol\leftarrow$ \textbf{Llama3-8B}}   \\
      \multicolumn{5}{c}{\textbf{Stepwise Attention}  $\boldsymbol\leftarrow$  \textbf{Llama3-8B}} \\
     \midrule
    DSS  &48.0  &36.1 &30.3  &38.1  \\
    MMIloss &47.0  &37.9 &30.7  &38.5 \\
     \rowcolor{gray!25}
    MoLSAKI  &\textbf{49.5} &\textbf{39.8} &\textbf{32.2} &\textbf{40.5} \\
    \midrule
      \multicolumn{5}{c}{\textbf{Rationale} $\boldsymbol\leftarrow$ \textbf{PaLM-540B}} \\
       \multicolumn{5}{c}{\textbf{Stepwise Attention}  $\boldsymbol\leftarrow$ \textbf{Llama3-8B}} \\
     \midrule
    DSS &43.0 &33.3 &33.0 &36.4\\
    MMIloss &42.0 &29.6 &\textbf{34.9} &35.5 \\
     \rowcolor{gray!25}
    MoLSAKI &\textbf{45.5} &\textbf{37.9} &34.5 &\textbf{39.3} \\
    \bottomrule
  \end{tabular}
  \caption{Performance comparison of MoLSAKI and baselines on reasoning datasets across different teacher model configurations.}
  \label{svamp+palm}
\end{table}
To systematically evaluate the robustness of MoLSAKI across diverse teacher model configurations, we conduct comparative experiments under two distinct scenarios: 1) Unified Configuration employing Llama3-8B as the sole teacher for both rationale generation and Stepwise Attention on Numerical Tokens extraction, and 2) Hybrid Configuration combining PaLM-540B's rationale generation (following DSS) with Llama3-8B's numerical attention patterns. 

Using GPT2-Large as the student model across both settings, our experimental results (Table~\ref{svamp+palm}) yield three principal observations: 
First, configuration analysis reveals that despite PaLM-540B's substantial parameter advantage (540B vs 8B), its inferior mathematical reasoning capability, as evidenced by official benchmark~\cite{chowdhery2023palm,grattafiori2024llama} comparisons, explains the performance gap between configurations. Second, MoLSAKI demonstrates consistent superiority over baseline methods in both configurations, achieving significant relative accuracy improvements of 5.1\% (Unified) and 7.9\% (Hybrid), thereby validating its teacher-agnostic knowledge integration capability. Third, the architecture's decoupled design enables practical deployment flexibility, allowing simultaneous utilization of black-box models (e.g., GPT-4.1, Gemini2.5) for high-quality rationale generation and white-box models (e.g., Llama3-8B) for numerical attention extraction - an innovative paradigm for heterogeneous knowledge distillation.
These findings collectively substantiate MoLSAKI's effectiveness in cross-configuration applications while proposing a novel framework for optimally leveraging diverse model capabilities in knowledge transfer scenarios.

\section{Conclusion}
We contribute a new perspective to improving CoT distillation, positing that the stepwise attention on critical tokens implicitly encodes essential reasoning cues inherent in large models. Building upon this perspective, we propose MoLSAKI, a novel distillation framework aimed at resolving the issue of critical information underutilization in CoT distillation for reasoning. It facilitates the transfer of the teacher model’s stepwise attention on critical tokens to the student model through a MoL strategy, which enables adaptive layer alignment.

\section*{Limitations}
Limited computational resources constrained our exploration of diverse model sizes and architectures for both teachers and students. Nevertheless, we believe this work offers a valuable perspective on Chain-of-Thought distillation and large language model reasoning. Despite its current scope, this study establishes a foundation for future research to extend attention-based distillation across a wider range of model scales and architectures. In addition, our experiments focused on relatively simple reasoning tasks with short chains of thought. Future work should examine whether these findings generalize to more complex problems that demand longer and more intricate reasoning paths.

\section*{Ethics Statement}
This research complies with established ethical standards. All datasets employed are publicly available and utilized solely for their intended research purposes. These datasets contain no personally identifiable information or sensitive content, thereby posing no risks to privacy or confidentiality. In addition, the study did not involve human subjects or annotators.

\section*{Acknowledgments}
We thank all anonymous reviewers for their insightful comments and suggestions. This work was supported by the National Natural Science Foundation of China (Grant No. 62572465).

\bibliography{custom}

@article{cobbe2021training,
  title={Training verifiers to solve math word problems},
  author={Cobbe, Karl and Kosaraju, Vineet and Bavarian, Mohammad and Chen, Mark and Jun, Heewoo and Kaiser, Lukasz and Plappert, Matthias and Tworek, Jerry and Hilton, Jacob and Nakano, Reiichiro and others},
  journal={arXiv preprint arXiv:2110.14168},
  year={2021}
}

@article{plaat2024reasoning,
  title={Reasoning with Large Language Models, a Survey},
  author={Plaat, Aske and Wong, Annie and Verberne, Suzan and Broekens, Joost and van Stein, Niki and Back, Thomas},
  journal={arXiv preprint arXiv:2407.11511},
  year={2024}
}

@inproceedings{chu-etal-2024-navigate,
    title = "Navigate through Enigmatic Labyrinth A Survey of Chain of Thought Reasoning: Advances, Frontiers and Future",
    author = "Chu, Zheng  and
      Chen, Jingchang  and
      Chen, Qianglong  and
      Yu, Weijiang  and
      He, Tao  and
      Wang, Haotian  and
      Peng, Weihua  and
      Liu, Ming  and
      Qin, Bing  and
      Liu, Ting",
    editor = "Ku, Lun-Wei  and
      Martins, Andre  and
      Srikumar, Vivek",
    booktitle = "Proceedings of the 62nd Annual Meeting of the Association for Computational Linguistics (Volume 1: Long Papers)",
    month = aug,
    year = "2024",
    address = "Bangkok, Thailand",
    publisher = "Association for Computational Linguistics",
    url = "https://aclanthology.org/2024.acl-long.65",
    pages = "1173--1203",
}

@article{liu2024aligning,
  title={Aligning teacher with student preferences for tailored training data generation},
  author={Liu, Yantao and Zhang, Zhao and Yao, Zijun and Cao, Shulin and Hou, Lei and Li, Juanzi},
  journal={arXiv preprint arXiv:2406.19227},
  year={2024}
}

@inproceedings{hsieh-etal-2023-distilling,
    title = "Distilling Step-by-Step! Outperforming Larger Language Models with Less Training Data and Smaller Model Sizes",
    author = "Hsieh, Cheng-Yu  and
      Li, Chun-Liang  and
      Yeh, Chih-kuan  and
      Nakhost, Hootan  and
      Fujii, Yasuhisa  and
      Ratner, Alex  and
      Krishna, Ranjay  and
      Lee, Chen-Yu  and
      Pfister, Tomas",
    editor = "Rogers, Anna  and
      Boyd-Graber, Jordan  and
      Okazaki, Naoaki",
    booktitle = "Findings of the Association for Computational Linguistics: ACL 2023",
    month = jul,
    year = "2023",
    address = "Toronto, Canada",
    publisher = "Association for Computational Linguistics",
    url = "https://aclanthology.org/2023.findings-acl.507",
    doi = "10.18653/v1/2023.findings-acl.507",
    pages = "8003--8017",
}

@inproceedings{chen-etal-2024-learning-maximize,
    title = "Learning to Maximize Mutual Information for Chain-of-Thought Distillation",
    author = "Chen, Xin  and
      Huang, Hanxian  and
      Gao, Yanjun  and
      Wang, Yi  and
      Zhao, Jishen  and
      Ding, Ke",
    editor = "Ku, Lun-Wei  and
      Martins, Andre  and
      Srikumar, Vivek",
    booktitle = "Findings of the Association for Computational Linguistics: ACL 2024",
    month = aug,
    year = "2024",
    address = "Bangkok, Thailand",
    publisher = "Association for Computational Linguistics",
    url = "https://aclanthology.org/2024.findings-acl.409",
    doi = "10.18653/v1/2024.findings-acl.409",
    pages = "6857--6868",
}

@inproceedings{zhang-etal-2024-dual,
    title = "Dual-Space Knowledge Distillation for Large Language Models",
    author = "Zhang, Songming  and
      Zhang, Xue  and
      Sun, Zengkui  and
      Chen, Yufeng  and
      Xu, Jinan",
    editor = "Al-Onaizan, Yaser  and
      Bansal, Mohit  and
      Chen, Yun-Nung",
    booktitle = "Proceedings of the 2024 Conference on Empirical Methods in Natural Language Processing",
    month = nov,
    year = "2024",
    address = "Miami, Florida, USA",
    publisher = "Association for Computational Linguistics",
    url = "https://aclanthology.org/2024.emnlp-main.1010",
    doi = "10.18653/v1/2024.emnlp-main.1010",
    pages = "18164--18181",
}

@inproceedings{ho-etal-2023-large,
    title = "Large Language Models Are Reasoning Teachers",
    author = "Ho, Namgyu  and
      Schmid, Laura  and
      Yun, Se-Young",
    editor = "Rogers, Anna  and
      Boyd-Graber, Jordan  and
      Okazaki, Naoaki",
    booktitle = "Proceedings of the 61st Annual Meeting of the Association for Computational Linguistics (Volume 1: Long Papers)",
    month = jul,
    year = "2023",
    address = "Toronto, Canada",
    publisher = "Association for Computational Linguistics",
    url = "https://aclanthology.org/2023.acl-long.830",
    doi = "10.18653/v1/2023.acl-long.830",
    pages = "14852--14882",
}

@inproceedings{fu2023specializing,
  title={Specializing smaller language models towards multi-step reasoning},
  author={Fu, Yao and Peng, Hao and Ou, Litu and Sabharwal, Ashish and Khot, Tushar},
  booktitle={International Conference on Machine Learning},
  pages={10421--10430},
  year={2023},
  organization={PMLR}
}

@inproceedings{li-etal-2023-symbolic,
    title = "Symbolic Chain-of-Thought Distillation: Small Models Can Also {``}Think{''} Step-by-Step",
    author = "Li, Liunian Harold  and
      Hessel, Jack  and
      Yu, Youngjae  and
      Ren, Xiang  and
      Chang, Kai-Wei  and
      Choi, Yejin",
    editor = "Rogers, Anna  and
      Boyd-Graber, Jordan  and
      Okazaki, Naoaki",
    booktitle = "Proceedings of the 61st Annual Meeting of the Association for Computational Linguistics (Volume 1: Long Papers)",
    month = jul,
    year = "2023",
    address = "Toronto, Canada",
    publisher = "Association for Computational Linguistics",
    url = "https://aclanthology.org/2023.acl-long.150",
    doi = "10.18653/v1/2023.acl-long.150",
    pages = "2665--2679",
}

@inproceedings{patel-etal-2021-nlp,
    title = "Are {NLP} Models really able to Solve Simple Math Word Problems?",
    author = "Patel, Arkil  and
      Bhattamishra, Satwik  and
      Goyal, Navin",
    editor = "Toutanova, Kristina  and
      Rumshisky, Anna  and
      Zettlemoyer, Luke  and
      Hakkani-Tur, Dilek  and
      Beltagy, Iz  and
      Bethard, Steven  and
      Cotterell, Ryan  and
      Chakraborty, Tanmoy  and
      Zhou, Yichao",
    booktitle = "Proceedings of the 2021 Conference of the North American Chapter of the Association for Computational Linguistics: Human Language Technologies",
    month = jun,
    year = "2021",
    address = "Online",
    publisher = "Association for Computational Linguistics",
    url = "https://aclanthology.org/2021.naacl-main.168/",
    doi = "10.18653/v1/2021.naacl-main.168",
    pages = "2080--2094"
}

@article{koncel-kedziorski-etal-2015-parsing,
    title = "Parsing Algebraic Word Problems into Equations",
    author = "Koncel-Kedziorski, Rik  and
      Hajishirzi, Hannaneh  and
      Sabharwal, Ashish  and
      Etzioni, Oren  and
      Ang, Siena Dumas",
    editor = "Collins, Michael  and
      Lee, Lillian",
    journal = "Transactions of the Association for Computational Linguistics",
    volume = "3",
    year = "2015",
    address = "Cambridge, MA",
    publisher = "MIT Press",
    url = "https://aclanthology.org/Q15-1042/",
    doi = "10.1162/tacl_a_00160",
    pages = "585--597"
}

@article{miao2021diverse,
  title={A diverse corpus for evaluating and developing English math word problem solvers},
  author={Miao, Shen-Yun and Liang, Chao-Chun and Su, Keh-Yih},
  journal={arXiv preprint arXiv:2106.15772},
  year={2021}
}

@article{kojima2022large,
  title={Large language models are zero-shot reasoners},
  author={Kojima, Takeshi and Gu, Shixiang Shane and Reid, Machel and Matsuo, Yutaka and Iwasawa, Yusuke},
  journal={Advances in neural information processing systems},
  volume={35},
  pages={22199--22213},
  year={2022}
}

@article{wei2022chain,
  title={Chain-of-thought prompting elicits reasoning in large language models},
  author={Wei, Jason and Wang, Xuezhi and Schuurmans, Dale and Bosma, Maarten and Xia, Fei and Chi, Ed and Le, Quoc V and Zhou, Denny and others},
  journal={Advances in neural information processing systems},
  volume={35},
  pages={24824--24837},
  year={2022}
}

@inproceedings{lee2024mentor,
  title={Mentor-KD: Making Small Language Models Better Multi-step Reasoners},
  author={Lee, Hojae and Kim, Junho and Lee, SangKeun},
  booktitle={Proceedings of the 2024 Conference on Empirical Methods in Natural Language Processing},
  pages={17643--17658},
  year={2024}
}

@inproceedings{jiao2020tinybert,
  title={TinyBERT: Distilling BERT for Natural Language Understanding},
  author={Jiao, Xiaoqi and Yin, Yichun and Shang, Lifeng and Jiang, Xin and Chen, Xiao and Li, Linlin and Wang, Fang and Liu, Qun},
  booktitle={Findings of the Association for Computational Linguistics: EMNLP 2020},
  pages={4163--4174},
  year={2020}
}

@inproceedings{sun2020mobilebert,
  title={MobileBERT: a Compact Task-Agnostic BERT for Resource-Limited Devices},
  author={Sun, Zhiqing and Yu, Hongkun and Song, Xiaodan and Liu, Renjie and Yang, Yiming and Zhou, Denny},
  booktitle={Proceedings of the 58th Annual Meeting of the Association for Computational Linguistics},
  pages={2158--2170},
  year={2020}
}

@article{wang2020minilm,
  title={Minilm: Deep self-attention distillation for task-agnostic compression of pre-trained transformers},
  author={Wang, Wenhui and Wei, Furu and Dong, Li and Bao, Hangbo and Yang, Nan and Zhou, Ming},
  journal={Advances in Neural Information Processing Systems},
  volume={33},
  pages={5776--5788},
  year={2020}
}

@article{xiao2023efficient,
  title={Efficient streaming language models with attention sinks},
  author={Xiao, Guangxuan and Tian, Yuandong and Chen, Beidi and Han, Song and Lewis, Mike},
  journal={arXiv preprint arXiv:2309.17453},
  year={2023}
}

@misc{Meta2024,
  author = {Meta},
  year = {2024},
  title = {Introducing Meta LLaMA 3: The Most Capable Openly Available LLM to Date},
  howpublished = {Blog},
  url = {https://example.com},
  note = {Accessed: 2024-01-24} 
}

@inproceedings{radford2019language,
  title={Language Models are Unsupervised Multitask Learners},
  author={Radford, Alec and Narasimhan, Karthik and Salimans, Tim and Sutskever, Ilya},
  booktitle={OpenAI Blog},
  year={2019},
  url={https://openai.com/research/language-unsupervised}
}

@article{chowdhery2023palm,
  title={Palm: Scaling language modeling with pathways},
  author={Chowdhery, Aakanksha and Narang, Sharan and Devlin, Jacob and Bosma, Maarten and Mishra, Gaurav and Roberts, Adam and Barham, Paul and Chung, Hyung Won and Sutton, Charles and Gehrmann, Sebastian and others},
  journal={Journal of Machine Learning Research},
  volume={24},
  number={240},
  pages={1--113},
  year={2023}
}

@article{grattafiori2024llama,
  title={The llama 3 herd of models},
  author={Grattafiori, Aaron and Dubey, Abhimanyu and Jauhri, Abhinav and Pandey, Abhinav and Kadian, Abhishek and Al-Dahle, Ahmad and Letman, Aiesha and Mathur, Akhil and Schelten, Alan and Vaughan, Alex and others},
  journal={arXiv e-prints},
  pages={arXiv--2407},
  year={2024}
}

@article{jin2024moh,
  title={Moh: Multi-head attention as mixture-of-head attention},
  author={Jin, Peng and Zhu, Bo and Yuan, Li and Yan, Shuicheng},
  journal={arXiv preprint arXiv:2410.11842},
  year={2024}
}

@article{zhou2022mixture,
  title={Mixture-of-experts with expert choice routing},
  author={Zhou, Yanqi and Lei, Tao and Liu, Hanxiao and Du, Nan and Huang, Yanping and Zhao, Vincent and Dai, Andrew M and Le, Quoc V and Laudon, James and others},
  journal={Advances in Neural Information Processing Systems},
  volume={35},
  pages={7103--7114},
  year={2022}
}

@article{zhang2019root,
  title={Root mean square layer normalization},
  author={Zhang, Biao and Sennrich, Rico},
  journal={Advances in Neural Information Processing Systems},
  volume={32},
  year={2019}
}

@article{hu2024minicpm,
  title={Minicpm: Unveiling the potential of small language models with scalable training strategies},
  author={Hu, Shengding and Tu, Yuge and Han, Xu and He, Chaoqun and Cui, Ganqu and Long, Xiang and Zheng, Zhi and Fang, Yewei and Huang, Yuxiang and Zhao, Weilin and others},
  journal={arXiv preprint arXiv:2404.06395},
  year={2024}
}

@article{barbero2025llms,
  title={Why do LLMs attend to the first token?},
  author={Barbero, Federico and Arroyo, Alvaro and Gu, Xiangming and Perivolaropoulos, Christos and Bronstein, Michael and Pascanu, Razvan and others},
  journal={arXiv preprint arXiv:2504.02732},
  year={2025}
}

@article{yao2024knowledge,
  title={Knowledge circuits in pretrained transformers},
  author={Yao, Yunzhi and Zhang, Ningyu and Xi, Zekun and Wang, Mengru and Xu, Ziwen and Deng, Shumin and Chen, Huajun},
  journal={arXiv preprint arXiv:2405.17969},
  year={2024}
}

@inproceedings{geva2023dissecting,
  title={Dissecting Recall of Factual Associations in Auto-Regressive Language Models},
  author={Geva, Mor and Bastings, Jasmijn and Filippova, Katja and Globerson, Amir},
  booktitle={Proceedings of the 2023 Conference on Empirical Methods in Natural Language Processing},
  pages={12216--12235},
  year={2023}
}

@article{zheng2024attention,
  title={Attention heads of large language models: A survey},
  author={Zheng, Zifan and Wang, Yezhaohui and Huang, Yuxin and Song, Shichao and Yang, Mingchuan and Tang, Bo and Xiong, Feiyu and Li, Zhiyu},
  journal={arXiv preprint arXiv:2409.03752},
  year={2024}
}

@inproceedings{talmor-etal-2019-commonsenseqa,
    title = "{C}ommonsense{QA}: A Question Answering Challenge Targeting Commonsense Knowledge",
    author = "Talmor, Alon  and
      Herzig, Jonathan  and
      Lourie, Nicholas  and
      Berant, Jonathan",
    editor = "Burstein, Jill  and
      Doran, Christy  and
      Solorio, Thamar",
    booktitle = "Proceedings of the 2019 Conference of the North {A}merican Chapter of the Association for Computational Linguistics: Human Language Technologies, Volume 1 (Long and Short Papers)",
    month = jun,
    year = "2019",
    address = "Minneapolis, Minnesota",
    publisher = "Association for Computational Linguistics",
    url = "https://aclanthology.org/N19-1421/",
    doi = "10.18653/v1/N19-1421",
    pages = "4149--4158"
}

@article{zhang2024tinyllama,
  title={Tinyllama: An open-source small language model},
  author={Zhang, Peiyuan and Zeng, Guangtao and Wang, Tianduo and Lu, Wei},
  journal={arXiv preprint arXiv:2401.02385},
  year={2024}
}

@misc{qwen2025qwen25technicalreport,
      title={Qwen2.5 Technical Report}, 
      author={Qwen and : and An Yang and Baosong Yang and Beichen Zhang and Binyuan Hui and Bo Zheng and Bowen Yu and Chengyuan Li and Dayiheng Liu and Fei Huang and Haoran Wei and Huan Lin and Jian Yang and Jianhong Tu and Jianwei Zhang and Jianxin Yang and Jiaxi Yang and Jingren Zhou and Junyang Lin and Kai Dang and Keming Lu and Keqin Bao and Kexin Yang and Le Yu and Mei Li and Mingfeng Xue and Pei Zhang and Qin Zhu and Rui Men and Runji Lin and Tianhao Li and Tianyi Tang and Tingyu Xia and Xingzhang Ren and Xuancheng Ren and Yang Fan and Yang Su and Yichang Zhang and Yu Wan and Yuqiong Liu and Zeyu Cui and Zhenru Zhang and Zihan Qiu},
      year={2025},
      eprint={2412.15115},
      archivePrefix={arXiv},
      primaryClass={cs.CL},
      url={https://arxiv.org/abs/2412.15115}, 
}

@article{zhang2025s1,
  title={S1-bench: A simple benchmark for evaluating system 1 thinking capability of large reasoning models},
  author={Zhang, Wenyuan and Nie, Shuaiyi and Zhang, Xinghua and Zhang, Zefeng and Liu, Tingwen},
  journal={arXiv preprint arXiv:2504.10368},
  year={2025}
}

@article{li2025system,
  title={From system 1 to system 2: A survey of reasoning large language models},
  author={Li, Zhong-Zhi and Zhang, Duzhen and Zhang, Ming-Liang and Zhang, Jiaxin and Liu, Zengyan and Yao, Yuxuan and Xu, Haotian and Zheng, Junhao and Wang, Pei-Jie and Chen, Xiuyi and others},
  journal={arXiv preprint arXiv:2502.17419},
  year={2025}
}

@inproceedings{chen-etal-2025-inner,
    title = "Inner Thinking Transformer: Leveraging Dynamic Depth Scaling to Foster Adaptive Internal Thinking",
    author = "Chen, Yilong  and
      Shang, Junyuan  and
      Zhang, Zhenyu  and
      Xie, Yanxi  and
      Sheng, Jiawei  and
      Liu, Tingwen  and
      Wang, Shuohuan  and
      Sun, Yu  and
      Wu, Hua  and
      Wang, Haifeng",
    editor = "Che, Wanxiang  and
      Nabende, Joyce  and
      Shutova, Ekaterina  and
      Pilehvar, Mohammad Taher",
    booktitle = "Proceedings of the 63rd Annual Meeting of the Association for Computational Linguistics (Volume 1: Long Papers)",
    month = jul,
    year = "2025",
    address = "Vienna, Austria",
    publisher = "Association for Computational Linguistics",
    url = "https://aclanthology.org/2025.acl-long.1369/",
    doi = "10.18653/v1/2025.acl-long.1369",
    pages = "28241--28259",
    ISBN = "979-8-89176-251-0"
}

@article{zhang2024revealing,
  title={Revealing and Mitigating the Challenge of Detecting Character Knowledge Errors in LLM Role-Playing},
  author={Zhang, Wenyuan and Nie, Shuaiyi and Sheng, Jiawei and Zhang, Zefeng and Zhang, Xinghua and He, Yongquan and Liu, Tingwen},
  journal={arXiv preprint arXiv:2409.11726},
  year={2024}
}

@inproceedings{zhang-etal-2025-sotopia,
    title = "{SOTOPIA}-{\ensuremath{\Omega}}: Dynamic Strategy Injection Learning and Social Instruction Following Evaluation for Social Agents",
    author = "Zhang, Wenyuan  and
      Liu, Tianyun  and
      Song, Mengxiao  and
      Li, Xiaodong  and
      Liu, Tingwen",
    editor = "Che, Wanxiang  and
      Nabende, Joyce  and
      Shutova, Ekaterina  and
      Pilehvar, Mohammad Taher",
    booktitle = "Proceedings of the 63rd Annual Meeting of the Association for Computational Linguistics (Volume 1: Long Papers)",
    month = jul,
    year = "2025",
    address = "Vienna, Austria",
    publisher = "Association for Computational Linguistics",
    url = "https://aclanthology.org/2025.acl-long.1203/",
    doi = "10.18653/v1/2025.acl-long.1203",
    pages = "24669--24697",
    ISBN = "979-8-89176-251-0"
}

\clearpage

\appendix

\section{Related Work}
\label{app:Related Work}

\subsection{Chain-of-Thought Distillation}
\label{subapp: Chain of Thought Distillation}
In complex reasoning, CoT distillation methods typically transfer the step-by-step rationales generated by the teacher model to the student model to enhance its reasoning abilities~\cite{ho-etal-2023-large, fu2023specializing,li-etal-2023-symbolic,hsieh-etal-2023-distilling,zhang2024revealing,zhang-etal-2025-sotopia}.
DSS \cite{hsieh-etal-2023-distilling} treats CoT distillation as a multitask learning problem, assigning two labels per query: the final answer and the rationale generated by the teacher model.
Following this, several studies incorporate an auxiliary loss to further enhance the complex reasoning capabilities of small language models.
Mentor-KD~\cite{lee2024mentor} introduces a mentor model situated between the student and teacher models with its logit output distribution that serves as an auxiliary soft label for distillation. 
MMIloss~\cite{chen-etal-2024-learning-maximize} introduces a maximum mutual information loss as an additional distilling objective, addressing DSS’s oversight of the mutual information between the rationale and final answer.

However, existing CoT distillation primarily focuses on transferring the result of the teacher's reasoning (the rationales), rather than the process itself. We contend that the teacher's ability to progressively attend to critical tokens during reasoning is a more fundamental and valuable skill. Hence, our goal is to transfer this crucial progressive attention pattern to the student model. We achieve this by incorporating a stepwise attention on critical tokens distillation loss $L_{att}$ (Eq.(\ref{Eq:loss_all})), which encourages the student to learn this vital ability to focus on key information step-by-step.

\subsection{Self-Attention Distillation}
\label{subapp: Self-Attention Distillation}
Existing self-attention distillation methods~\cite{jiao2020tinybert,sun2020mobilebert,wang2020minilm} suffer from two main drawbacks. They are not designed for reasoning tasks, often neglecting reasoning-specific attention patterns and distilling the full self-attention matrix, which mandates identical teacher-student tokenizers. Moreover, they typically handle varying teacher-student layer counts with rigid single-layer alignment (SL), ignoring the functional diversity of layers~\cite{geva2023dissecting,yao2024knowledge}. Conversely, we utilize a more flexible MoL layer alignment strategy, whose superiority over SL is demonstrated in our ablation studies (see in Sec.\textbf{\ref{Sec: Fixed Layer Alignment vs Adaptive Layer Alignment}}).

\section{Critical Tokens in CoT}
\label{app:Critical Tokens in CoT}
\citet{chen-etal-2025-inner} investigated critical tokens during the pre-training stage and allocated more computational resources to these tokens, whereas our work primarily focuses on critical tokens within the chain-of-thought (CoT) during the reasoning process.
Effective reasoning relies on focusing attention on critical information, which provides essential clues for successful problem-solving, much like in human cognition. Driven by this understanding, we sought to analyse the attention distribution over critical tokens in LLMs' CoT.
Although prior research~\cite{xiao2023efficient,barbero2025llms} indicates that autoregressive LLMs often prioritize the initial token, our specific interest lies in the attention distribution across the remaining tokens within the reasoning steps. 
To highlight this, we developed methods tailored to mathematical and commonsense reasoning to identify these critical tokens and performed visualization analysis, omitting the initial token's attention.

\subsection{Mathematical Reasoning}
\label{app: Mathematical Reasoning}
Recognizing the intuitive importance of numerical tokens in mathematical reasoning, we conducted a visualization analysis to confirm this. 
We conducted analysis experiments on the mathematical reasoning datasets GSM8K and SVAMP (100 randomly sampled instances from each, totaling 200 samples). 

The student model was fine-tuned using the DSS~\cite{hsieh-etal-2023-distilling} method on CoT data from GSM8K and SVAMP to endow it with mathematical reasoning capability. 
For each sample, given the question and the zero-shot prompt 'Let's think step by step,' the model was tasked with generating a reasoning process. Attention scores for numerical and non-numerical tokens were recorded during this process. 
To compute the step-by-step attention, we segmented the reasoning steps based on periods and applied an averaging strategy across layers and attention heads. 
Finally, we applied the softmax function to the average attention scores of numerical and non-numerical tokens, yielding the relative proportions of stepwise attention allocation (see Figure {\ref{fig:intro3}}).

The results demonstrate that both the teacher and student models allocate higher average attention weights to numerical tokens compared to non-numerical tokens in the question during mathematical reasoning. 
These findings collectively underscore the critical role of numerical tokens in mathematical reasoning.

\subsection{Commonsense Reasoning}
\label{app: Commense Reasoning}
\begin{figure}[!tb] 
    \centering
    \includegraphics[width=\linewidth]{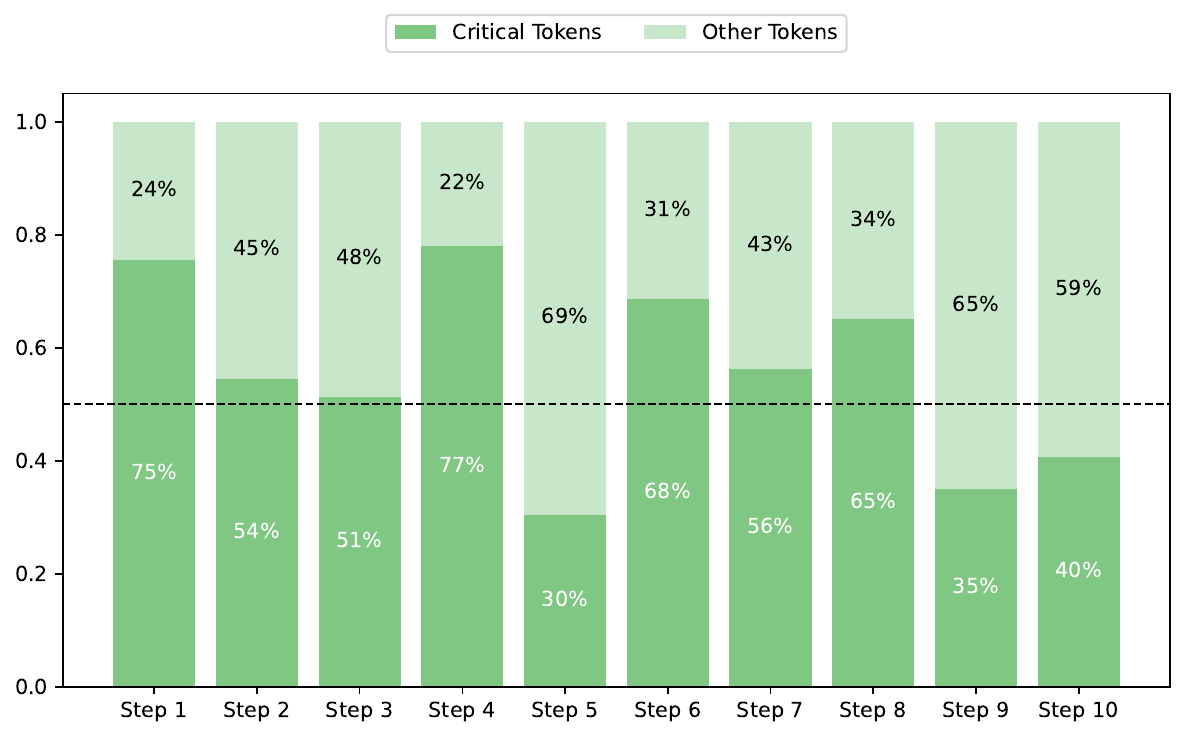} 
        \caption{We randomly sampled 100 instances from CommonSenseQA to analyze the average attention allocated by Qwen2.5-32B to critical tokens corresponding to keywords relative to other tokens at each step.} \label{fig: qwen1}
    \label{fig: qwen_portion}
\end{figure}

For commonsense reasoning, we adopt a keyword extraction method. 
To obtain critical tokens using this method, we design a prompt during few-shot CoT generation by the teacher model, asking it to list 3-8 unique keywords deemed crucial for reasoning (in Figure {\ref{prompt3}}). 
The visualization analysis method was the same as described in Appendix~\ref{app: Mathematical Reasoning}. 
Our results indicate that, in most steps of the inference process, tokens corresponding to extracted keywords received significantly higher attention, highlighting their crucial role in facilitating inference. (see Figure \ref{fig: qwen_portion}).

\section{Extracting Stepwise Attention}

\subsection{Extracting Indices of Critical Tokens}
\label{app: Extract Indices of Critical Tokens}
Critical tokens' indices are automatically extracted from the token sequence, eliminating the need for manual annotation. 
We first identify critical words based on the task type: 
For commonsense reasoning, we prompt the teacher model during CoT generation to provide keywords it deemed important for reasoning.
For mathematical reasoning, numerical words are considered critical based on prior analysis. This is achieved by first locating numerical words or teacher-provided keywords in the text sequence via regex matching.

Once these critical words are identified in the input text sequence, we automatically extract their corresponding indices ($\mathcal{M}_2$ in Eq.(\ref{Eq:cal_sa_t}) \& $\mathcal{N}_2$ in Eq.(\ref{Eq:cal_sa_s})) in the token sequence using regex matching and tokenizer mapping. 
It is noted that the elements within $\mathcal{M}_2$ and $\mathcal{N}_2$ are not simple integer indices, but rather are composed of smaller index sets.
Because a single critical word sometimes corresponds to multiple tokens, we obtain a small index set $\mathcal{P}\in\mathcal{M}_2$ ( $\mathcal{O}\in\mathcal{N}_2$) for each keyword.
In our stepwise attention calculation, these critical tokens in $\mathcal{P}$ are treated as a whole, and the attention received by all member tokens is summed to represent the attention received by the critical word. 

\subsection{Identical Shape}
\label{app: Identical Shape}
Notably, during student model distillation, the input CoT text and the critical words within the CoT are all derived from the teacher model. 
Consequently, the identical input text and critical words used in CoT distillation (see Figure {\ref{Framework}}) ensure that the stepwise attention matrix shares the same shape  ($|\mathcal{N}_1|=|\mathcal{M}_1|,|\mathcal{N}_2|=|\mathcal{M}_2|$) for teacher and student models, despite the significant difference between their tokenizers resulting in $\mathcal{N}_1\neq\mathcal{M}_1,\ \mathcal{N}_2\neq\mathcal{M}_2$.

\section{Progressive Attention Pattern}
\label{app: Progressive Attention Pattern on Critical Tokens}
Analogous to human reasoning, where attention to different key information shifts dynamically as steps evolve, we observe a dynamic pattern in the teacher model's stepwise attention towards critical tokens. 
This attention pattern implicitly encodes the teacher model's capture and utilization of key information during the reasoning process. 
To illustrate this, we visualized the stepwise attention on critical tokens from a specific layer of the teacher model for a reasoning sample. 
The results revealed distinct step-by-step variations in the teacher model's attention to critical tokens during reasoning (see Figure {\ref{fig:intro2}} \& Figure \ref{fig: qwen_layer}).

\section{Temperature Parameters}
\label{app: Temperature Parameters}
The temperature parameters $\tau_1$, $\tau_2$ control the sharpness of the adaptive layer weight distributions in the teacher and student models, respectively. The rationale for the MoL configuration and the specific values of hyperparameters $\tau_1=0.1$, $\tau_2=0.5$ is elaborated below.

\subsection{Teacher: $\mathbf{\tau_1}$}
\label{app: Teacher}
We conduct a visual analysis of the stepwise attention of the teacher model from both qualitative and quantitative perspectives.

\begin{figure*}[!htb]
  \includegraphics[width=\linewidth]{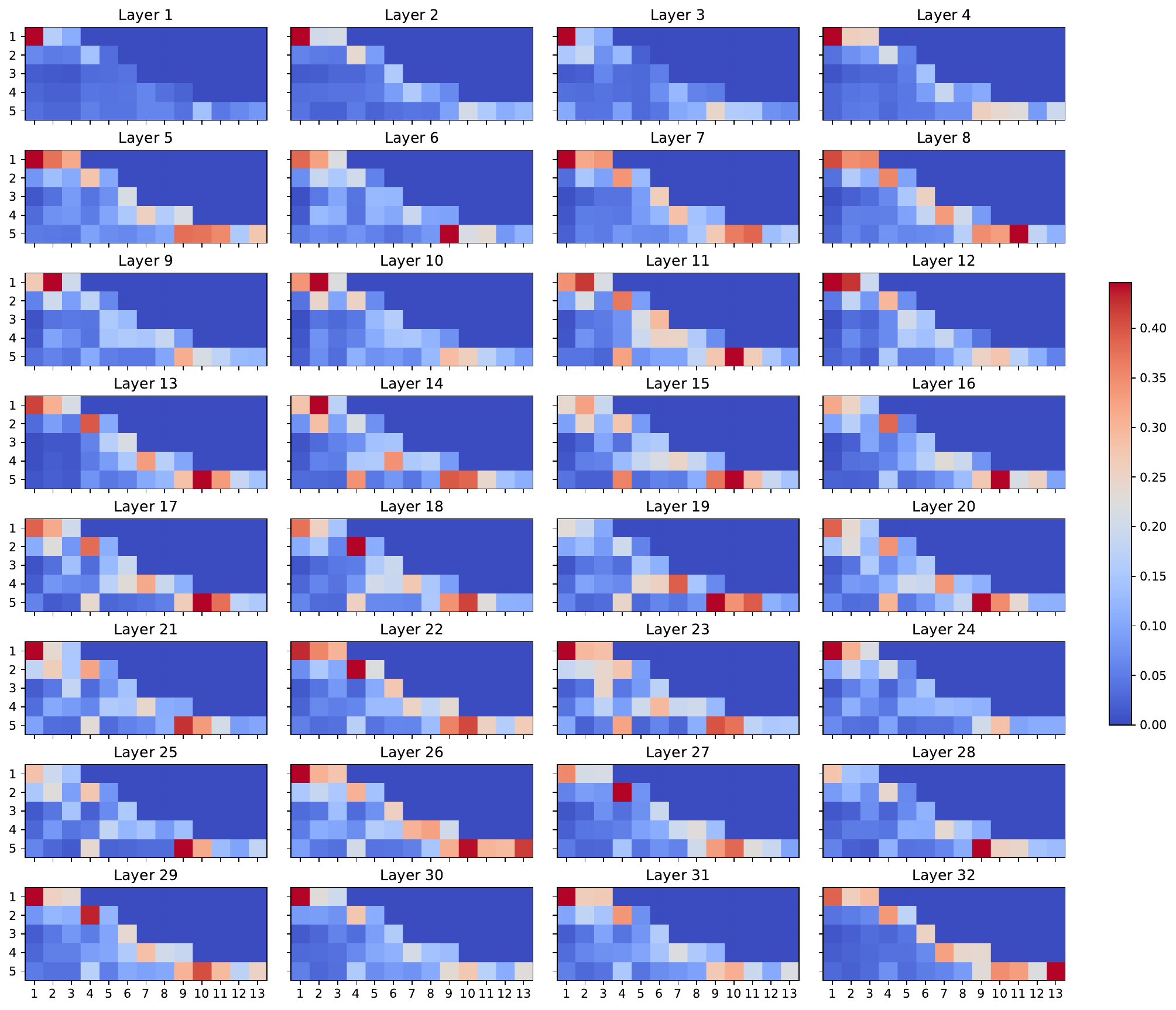} 
  \caption {\textbf{Stepwise attention heatmap on critical tokens from the teacher model Llama3-8B, which consists of 32 layers for a specific example.} For each layer, the average attention is computed from all attention heads. The horizontal axis represents the order of critical tokens, while the vertical axis indicates the step number (counted from the beginning of the question). The example is as follows: "Question: A mailman is tasked with delivering \textcolor[HTML]{b70b0f}{4} pieces of junk mail to each house in \textcolor[HTML]{b70b0f}{16} blocks, with each block containing \textcolor[HTML]{b70b0f}{17} houses. How many pieces of junk mail should he deliver in total? Rationale: The mailman delivers \textcolor[HTML]{b70b0f}{4} pieces of junk mail to each house in \textcolor[HTML]{b70b0f}{16} blocks, with each block containing \textcolor[HTML]{b70b0f}{17} houses. Therefore, the total number of houses is \textcolor[HTML]{b70b0f}{16} × \textcolor[HTML]{b70b0f}{17} = \textcolor[HTML]{b70b0f}{272}. Since the mailman delivers \textcolor[HTML]{b70b0f}{4} pieces of junk mail to each house, the total number of junk mail pieces is \textcolor[HTML]{b70b0f}{272} × \textcolor[HTML]{b70b0f}{4} = \textcolor[HTML]{b70b0f}{1088}."}
  \label{teacher_layer_attention}
\end{figure*}

\textbf{Qualitative:} We visualized the stepwise attention on critical tokens for a selected sample across all 32 layers of the teacher Llama3-8B model. As shown in Figure {\ref{teacher_layer_attention}}, this attention exhibits clear variations across different layers.

\textbf{Quantitative:} We randomly selected 100 samples from each of the GSM8K and SVAMP datasets and input the corresponding questions and rationales into the teacher Llama3-8B. We then extracted stepwise attention on critical tokens from each layer and computed their column gradients. Importantly, these column gradients are distinct from backpropagation gradients. Calculated using Eq.(\ref{eq:gradients}), they evaluate the average magnitude of attention weight differences between adjacent critical tokens. The column gradients reflect the magnitude of stepwise change in attention on critical tokens within a given layer. 
The results highlight significant attention shifts in the intermediate layers (with the most notable changes occurring in layers 13–15 in Figure {~\ref{fig:teacher_gradient}}). 

Through qualitative and quantitative analysis, we find that the most significant gradual change in attention to critical tokens takes place in the intermediate layers. This finding aligns with previous interpretability studies~\cite{geva2023dissecting,yao2024knowledge,zheng2024attention}, suggesting that the intermediate layers of large models are more strongly associated with reasoning than other layers. Aiming to ensure the student model prioritizes learning from these crucial intermediate layer patterns, we set $\tau_1$ to an extremely small value, thereby allocating them greater weight.

\subsection{Student: $\mathbf{\tau_2}$}
\label{app:student}
Since the student model is small, we aim for all its layers to participate substantially yet non-uniformly in attention distillation. To achieve this, $\tau_2$ is set to a moderate value, yielding a less peaked but non-uniform layer weight distribution.

\section{Details of Experiments}

\subsection{Datasets}
\label{Dataset Statistics}
The reasoning abilities of current large language models are generally categorized into two modes: System 1 thinking~\cite{zhang2025s1} and System 2 thinking~\cite{li2025system}.
In this work, we primarily focus on System 2 thinking.
To comprehensively evaluate performance across varying difficulty levels, we conducted experiments on five benchmarks spanning commonsense and mathematical reasoning. Table~\ref{table: dataset} provides data statistics for these benchmarks.

The mathematical reasoning datasets, all human-authored, consist primarily of grade school math word problems. Among these, GSM8K represents a challenging problem domain. While Asdiv was originally a multiple-choice mathematical reasoning dataset, we modified it by removing the options and rephrasing the questions as open-ended. This change was implemented to enhance task difficulty and minimize potential interference from random guessing.

For the commonsense reasoning benchmarks, Commonsense QA assesses the ability to apply everyday knowledge and commonsense reasoning about the physical and social world to answer questions in practical scenarios.

\begin{table}[!htb]
\centering
\small
\begin{tabular}{c|c|c|c}
\toprule
\textbf{Dataset} &\textbf{In-Domain} &\textbf{Train} &\textbf{Test} \\
\midrule
SVAMP  &\ding{51} & 800 & 200  \\
SingleEq  &\ding{55} & - & 108 \\
Asdiv  &\ding{55} & - & 406 \\
GSM8K  &\ding{55} & - & 1318  \\
CommonSenseQA  &\ding{51} & 9741 & 1221 \\
\bottomrule
\end{tabular}
\caption{Dataset statistics used in our experiments.}
\label{table: dataset}
\end{table}

\subsection{Hyperparameter Settings}
\label{Hyperparameter Settings}
All experiments were performed on the NVIDIA A800 $\times 1$ GPU cloud environments. The GPT2-Medium and GPT2-Large models were trained with the following configurations: learning rate = $5\times10^{-5}$, batch size = 16, maximum training steps = 4,000. The TinyLllama model was trained with the following configurations: learning rate = $1\times10^{-4}$, batch size = 16, maximum training steps = 2,000. We report the average results over three random runs.

\subsection{Prompts}
\label{app: Prompts}
\begin{figure*}[!htb] 
    \centering
    \begin{subfigure}[b]{0.95\textwidth} 
        \centering
        \includegraphics[width=0.95\textwidth]{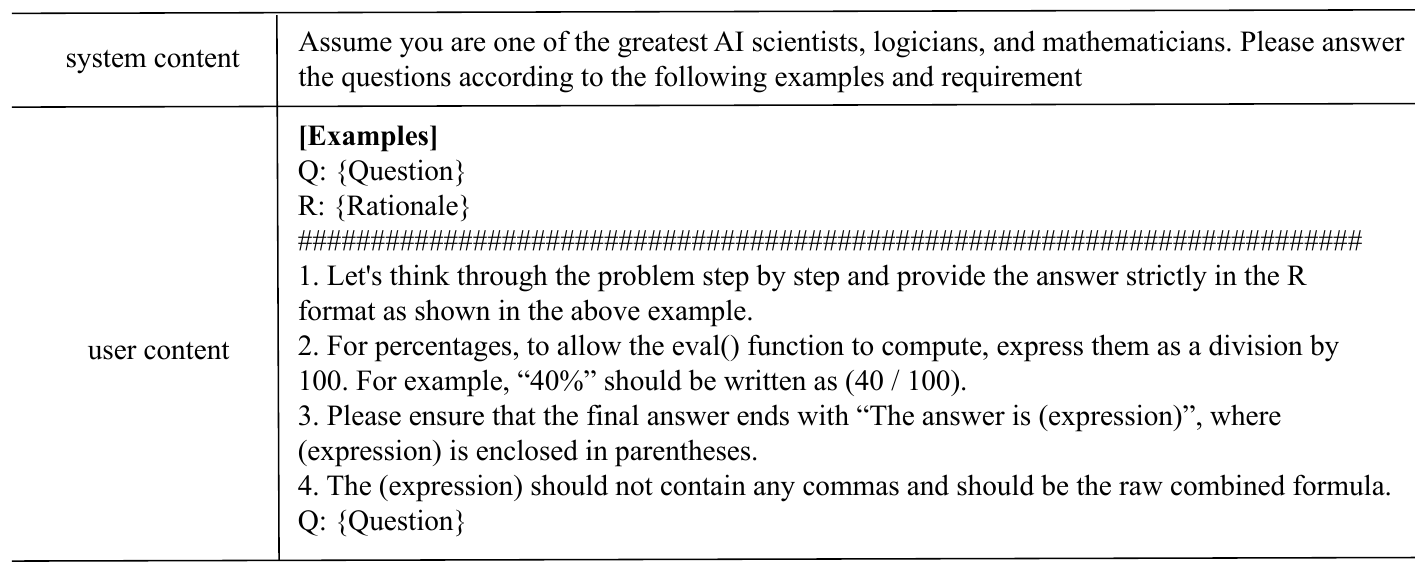} 
        \caption{Generating CoT when given the question.}
        \label{prompt1}
    \end{subfigure}
        
    \begin{subfigure}[b]{0.95\textwidth}
        \centering
        \includegraphics[width=0.95\textwidth]{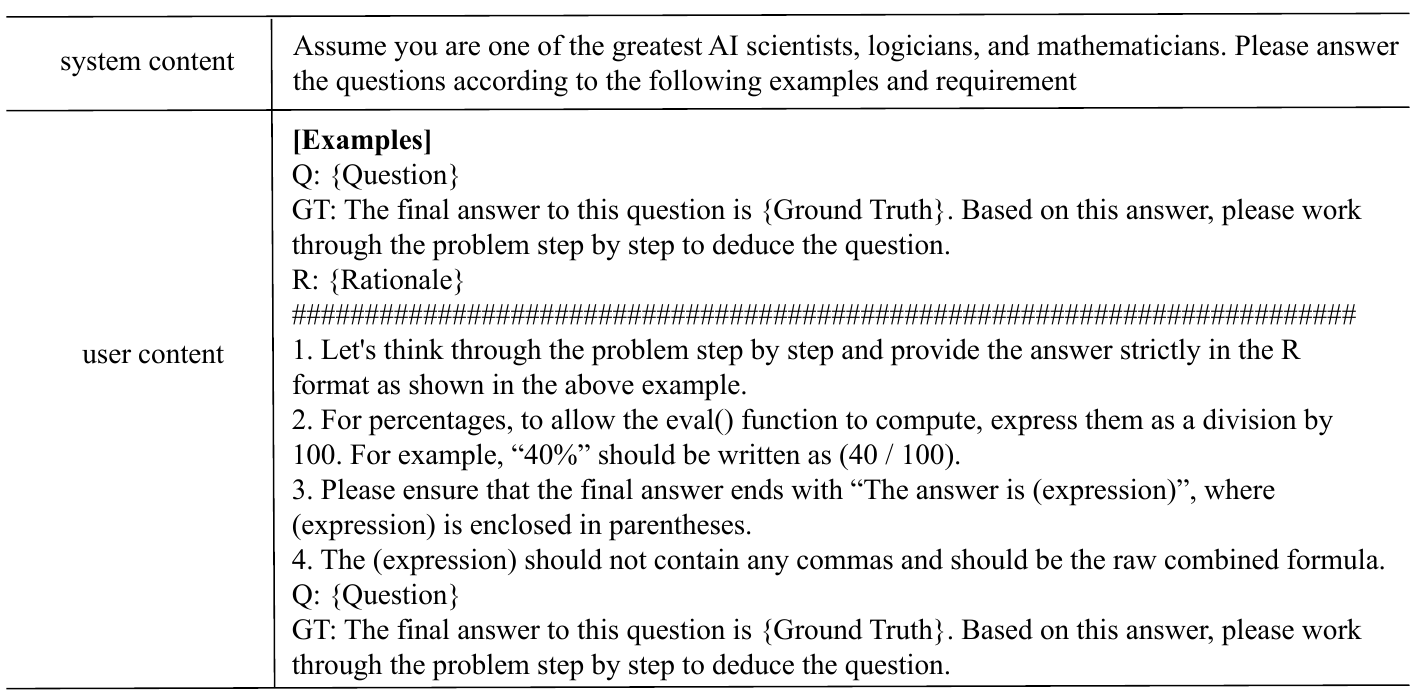}
       \caption{Generating CoT when given the question and ground truth.}
        \label{prompt2}
    \end{subfigure}
    \caption{Prompt template for generating CoT of the teacher model with dual-phase pipeline.}
    \label{prompt_template}
\end{figure*}

To obtain more accurate CoT samples, we design a dual-phase CoT generation pipeline to handle complexity-stratified questions. In the first stage, we prompt the teacher model to generate $r$ and $a$ based on the question $q$, similar to previous works. This initial phase ensures accurate responses for most relatively simple questions. The second stage addresses incorrect samples with higher complexity levels by prompting the teacher model to regenerate rationale $r$ and answer $\hat{a}$ under the guidance of both the question $q$ and the ground-truth $\hat{a}$. This dual-phase pipeline enables scalable generation of high-quality CoT samples while maintaining rigorous quality control throughout the process (details in Figure \ref{prompt1} \& Figure \ref{prompt2}).

\subsection{Layer Weight Visualization}
\label{Layer Weight Visualization}
\begin{figure}[!htb]
\includegraphics[width=\linewidth]{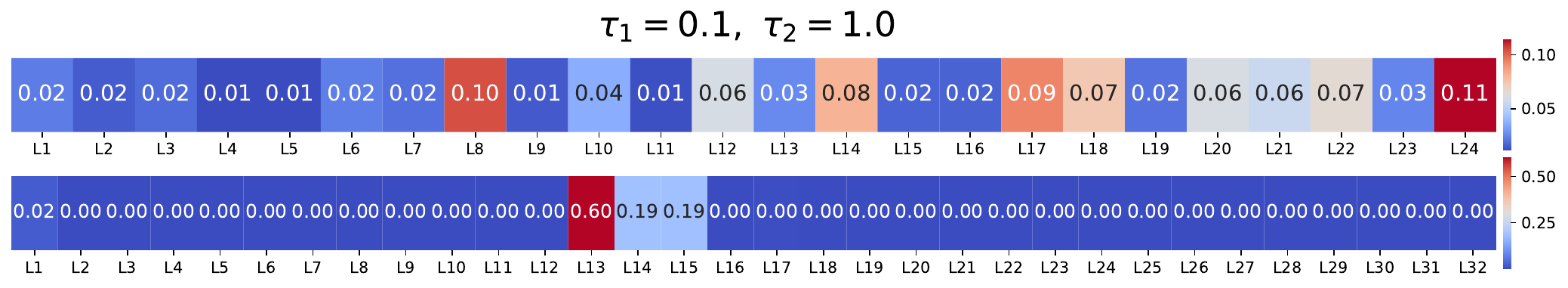} 
  \caption {Layer weight visualization when $\tau_2=1.0,\ \tau_1=0.1$ .}
  \label{s_t_weight_combined_4}
\end{figure}

We visualize the layer weights for the teacher and student models under $\tau_1$ = 0.1 and $\tau_2$ = 1.0, as shown in Figure~\ref{s_t_weight_combined_4}. Under this temperature parameter configuration, the weights of each layer in the student model are more evenly distributed. 

\subsection{Computational Cost}
\label{app: Computational Cost}
Regarding the computational overhead of MoLSAKI, the attention matrix is directly utilized as an intermediate result from the standard forward pass, thus introducing no additional computation. The newly introduced MoL module contributes only a marginal computational cost, consisting of one linear layer and RMSNorm. Consequently, the total increase in FLOPs from integrating MoLSAKI is slight, yielding significant distillation performance gains. To quantify this, we compared the FLOPs of all evaluated methods on the two student models.

\begin{table}[htb]
\centering
\small
\begin{tabular}{c|c|c}
\toprule
\textbf{Student Model} & \textbf{Method} & \textbf{FLOPs ($\times 10^{11}$)} \\ 
\midrule
\multirow{4}{*}{GPT2-Medium} & Vanilla Finetune & 0.947 \\
&DSS & 2.927 \\
&MMIloss & 2.9270010 \\ 
&MoLSAKI & 2.9270186 \\ 
\cmidrule{1 - 3}
\multirow{4}{*}{GPT2-Large} & Vanilla Finetune & 2.007 \\
&DSS & 6.081 \\
&MMIloss & 6.0810010 \\ 
&MoLSAKI & 6.0810273 \\ 
\bottomrule
\end{tabular}
\caption{Computational cost (FLOPs).}
\end{table}

\subsection{Case Study}
\label{app: Case Study}
\begin{figure*}[!htb] 
    \centering
    \begin{subfigure}[b]{0.95\textwidth} 
        \centering
        \includegraphics[width=0.95\textwidth]{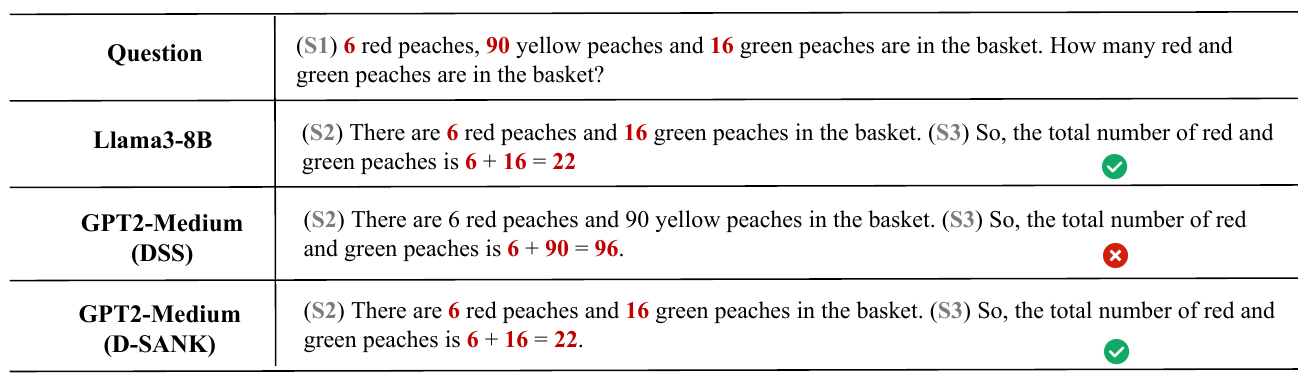} 
        \caption{}
        \label{case1}
    \end{subfigure}
        
    \begin{subfigure}[b]{0.95\textwidth}
        \centering
        \includegraphics[width=0.95\textwidth]{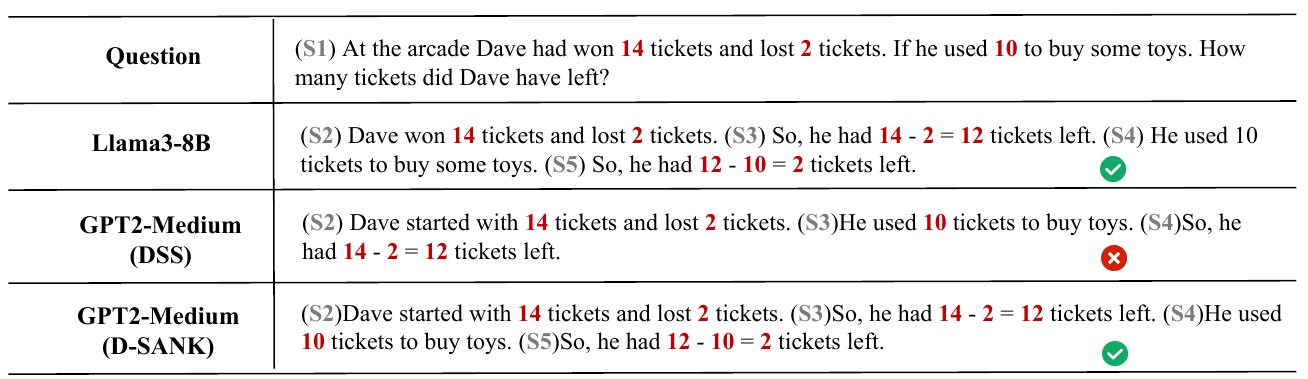}
        \caption{}
        \label{case2}
    \end{subfigure}

       \begin{subfigure}[b]{0.95\textwidth}
        \centering
        \includegraphics[width=0.95\textwidth]{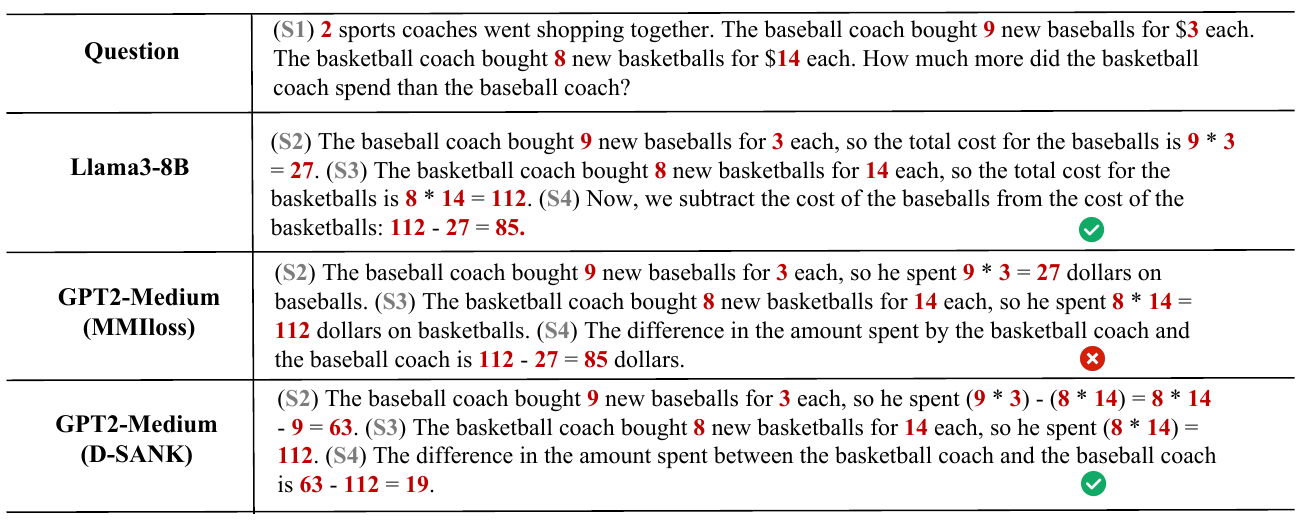}
        \caption{}
        \label{case3}
    \end{subfigure}

       \begin{subfigure}[b]{0.95\textwidth}
        \centering
        \includegraphics[width=0.95\textwidth]{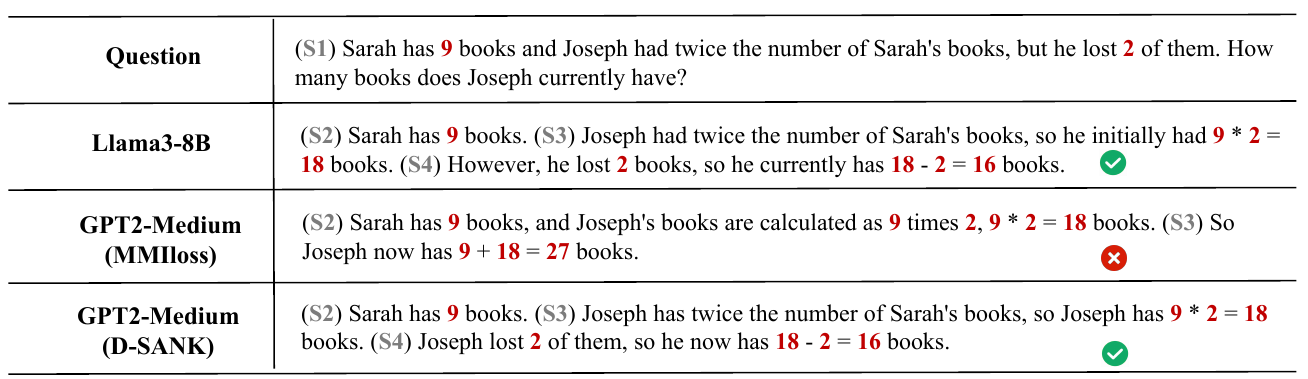}
        \caption{}
        \label{case4}
    \end{subfigure}
    \caption{Cases from SVAMP and GSM8K.}
    \label{case_study}
\end{figure*}

In this section, we select two samples each from SVAMP and GSM8K for case analysis (Figure~\ref{case_study}). We compare the differences between the MoLSAKI method and the baseline method in terms of rationale generation. And to further compare their stepwise attention on numerical tokens, we visualize the stepwise attention of the student model GPT2-Medium (8th layer) distilled by different methods and the teacher model Llama3-8B (13th layer).

When reasoning the question presented in Figure~\ref{case2}, the student model distilled by DSS mentions the condition "He used 10 tickets to buy toys" during the generation of rationales. However, it fails to utilize the number "10" in the subsequent reasoning process, leading to an incorrect result. When addressing the question in Figure~\ref{case4}, MMIloss overlooks the condition "but he lost 2 of them" while generating rationales, which also results in an incorrect answer. In contrast, the MoLSAKI method makes full and effective use of all relevant numerical conditions.

\begin{figure*}[!htb] 
    \centering
    \begin{subfigure}[b]{\textwidth} 
        \centering
        \includegraphics[width=\textwidth]{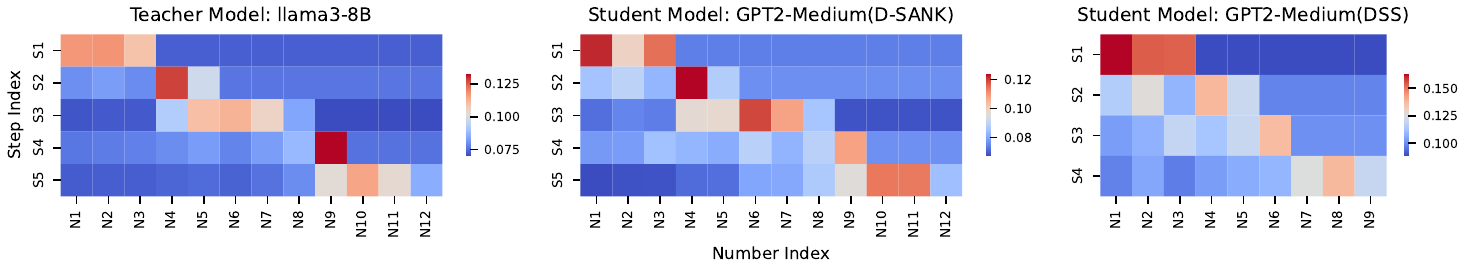} 
        \caption{The sample from the SVAMP dataset.}
        \label{case2_visual}
    \end{subfigure}
        
    \begin{subfigure}[b]{\textwidth}
        \centering
        \includegraphics[width=\textwidth]{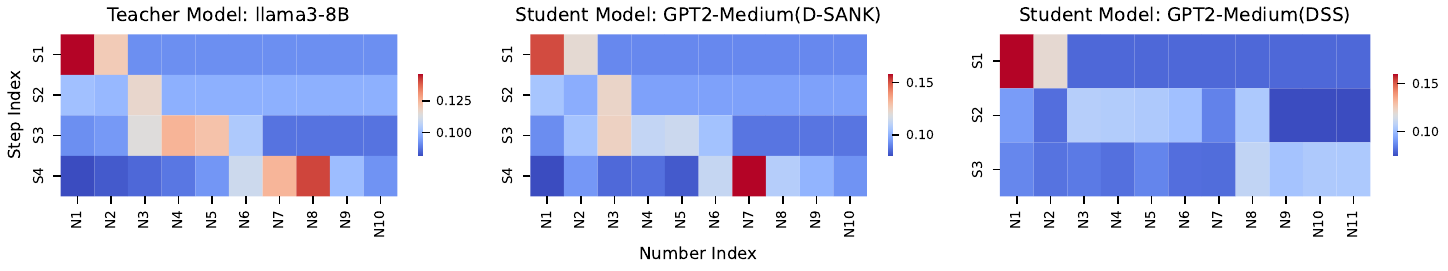}
        \caption{The sample from GSM8K dataset.}
        \label{case4_visual}
    \end{subfigure}
    \caption{We select one example each from SVAMP and GSM8K, visualizing stepwise attention on numerical tokens for student models distilled by DSS, MMIloss, and MoLSAKI, compared with the teacher model. Vertical and horizontal axes, respectively, denote the index of the step and the number.}
    \label{case_study_visual}
\end{figure*}

By comparing the stepwise attention on numerical tokens during the generation of rationales for the questions in the above two examples between the teacher model and the student models distilled by different methods in Figure~\ref{case_study_visual}, it can be observed that, compared with the baseline methods, the student model distilled by D-SANK exhibits a high degree of similarity to the teacher model in terms of stepwise attention. This indicates that distilling the teacher model's stepwise attention on number tokens to the student model can enhance the student model's comprehensive attention and in-depth understanding of numerical conditions. Consequently, the mathematical reasoning ability of the student model is improved.

\begin{figure*}[!htb]
        \centering
        \includegraphics[width=0.95\textwidth]{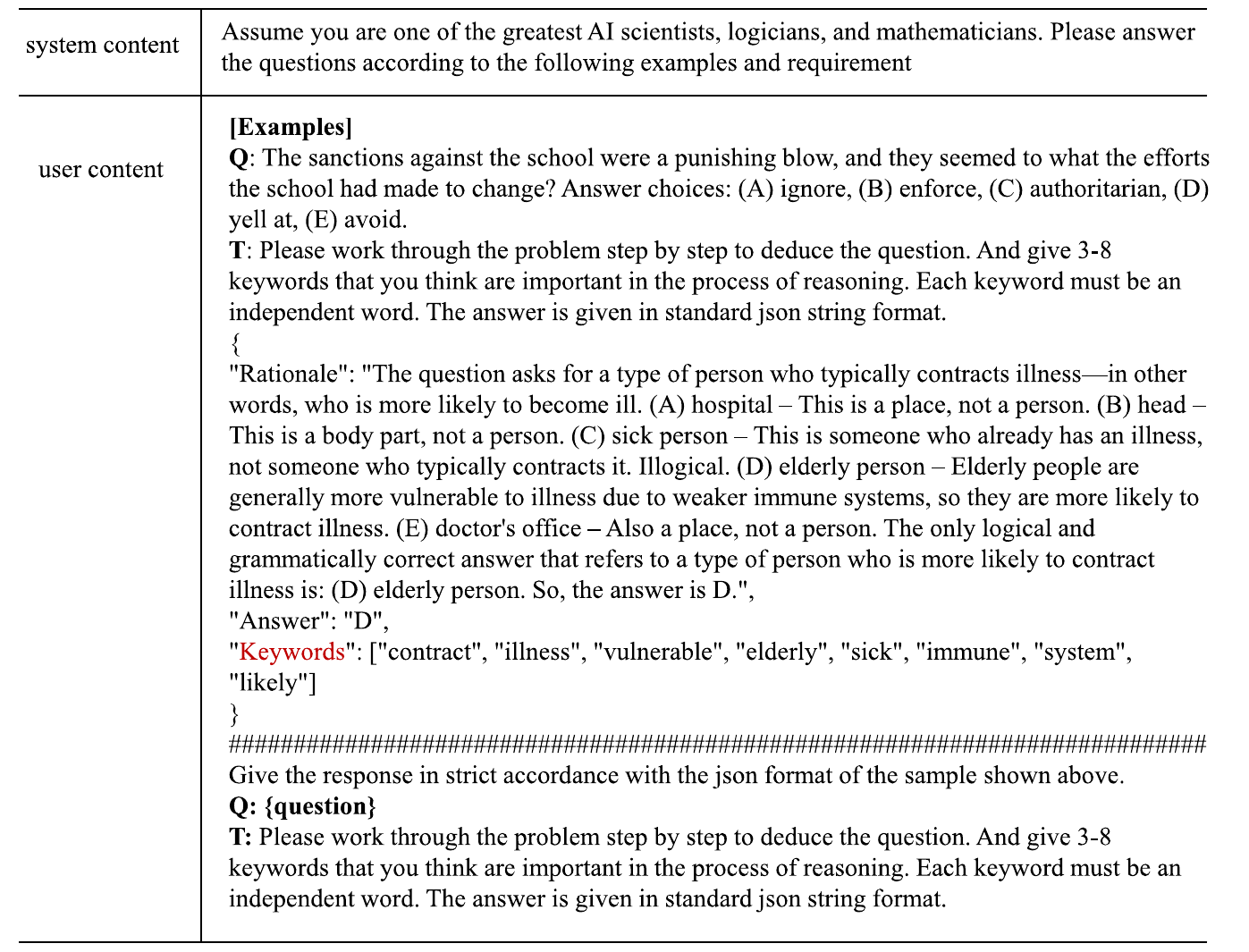} 
        \caption{Prompt template for generating keywords in the reasoning process of the teacher model.}
        \label{prompt3}
\end{figure*}

\end{document}